\documentclass[11pt]{article}
\usepackage{graphicx,amsmath,amsfonts,amssymb,fullpage,color}
\usepackage[natbibapa]{apacite}
\usepackage{caption}
\usepackage{subcaption}
\usepackage{setspace}
\usepackage{hyperref}
\usepackage{fancyhdr}
\title{\textbf{Building Machines That Learn and Think Like People}}
\date{}
\author{Brenden M. Lake,$^1$ Tomer D. Ullman,$^{2,4}$ Joshua B. Tenenbaum,$^{2,4}$ and Samuel J. Gershman$^{3,4}$ \\
$^1$Center for Data Science, New York University \\
$^2$Department of Brain and Cognitive Sciences, MIT \\
$^3$Department of Psychology and Center for Brain Science, Harvard University \\
$^4$Center for Brains Minds and Machines
}

\fancypagestyle{alim}{\fancyhf{}\fancyhead[L]{In press at \emph{Behavioral and Brain Sciences}.}}

\begin{document}
\maketitle

\thispagestyle{alim}

\begin{abstract}
Recent progress in artificial intelligence (AI) has renewed interest in building systems that learn and think like people. Many advances have come from using deep neural networks trained end-to-end in tasks such as object recognition, video games, and board games, achieving performance that equals or even beats humans in some respects. Despite their biological inspiration and performance achievements, these systems differ from human intelligence in crucial ways. We review progress in cognitive science suggesting that truly human-like learning and thinking machines will have to reach beyond current engineering trends in both what they learn, and how they learn it. Specifically, we argue that these machines should (a) build causal models of the world that support explanation and understanding, rather than merely solving pattern recognition problems; (b) ground learning in intuitive theories of physics and psychology, to support and enrich the knowledge that is learned; and (c) harness compositionality and learning-to-learn to rapidly acquire and generalize knowledge to new tasks and situations. We suggest concrete challenges and promising routes towards these goals that can combine the strengths of recent neural network advances with more structured cognitive models.
\end{abstract}

\section{Introduction}

Artificial intelligence (AI) has been a story of booms and busts, yet by any traditional measure of success, the last few years have been marked by exceptional progress. Much of this progress has come from recent advances in ``deep learning,'' characterized by learning large neural-network-style models with multiple layers of representation. These models have achieved remarkable gains in many domains spanning object recognition, speech recognition, and control \citep{Lecun2015,schmidhuber15}. In object recognition, \citet{Krizhevsky2012} trained a deep convolutional neural network \citep[convnets;][]{LeCun1989} that nearly halved the error rate of the previous state-of-the-art on the most challenging benchmark to date. In the years since, convnets continue to dominate, recently approaching human-level performance on some object recognition benchmarks \citep{Szegedy2014,Russakovsky2014,He2015}. In automatic speech recognition, Hidden Markov Models (HMMs) have been the leading approach since the late 1980s \citep{Juang1990}, yet this framework has been chipped away piece by piece and replaced with deep learning components \citep{Hinton2012}. Now, the leading approaches to speech recognition are fully neural network systems \citep{graves2013b,weng2014}. Ideas from deep learning have also been applied to learning complex control problems. \citet{Mnih2015} combined ideas from deep learning and reinforcement learning to make a ``deep reinforcement learning'' algorithm that learns to play large classes of simple video games from just frames of pixels and the game score, achieving human or superhuman level performance on many of these games \citep[see also][]{guo14,Stadie2015,Schaul2015}.

These accomplishments have helped neural networks regain their status as a leading paradigm in machine learning, much as they were in the late 1980s and early 1990s. The recent success of neural networks has captured attention beyond academia. In industry, companies such as Google and Facebook have active research divisions exploring these technologies, and object and speech recognition systems based on deep learning have been deployed in core products on smart phones and the web. The media has also covered many of the recent achievements of neural networks, often expressing the view that neural networks have achieved this recent success by virtue of their brain-like computation and thus their ability to emulate human learning and human cognition.

In this article, we view this excitement as an opportunity to examine what it means for a machine to learn or think like a person. We first review some of the criteria previously offered by cognitive scientists, developmental psychologists, and AI researchers. Second, we articulate what we view as the essential ingredients for building such a machine that learns or thinks like a person, synthesizing theoretical ideas and experimental data from research in cognitive science. Third, we consider contemporary AI (and deep learning in particular) in light of these ingredients, finding that deep learning models have yet to incorporate many of them and so may be solving some problems in different ways than people do. We end by discussing what we view as the most plausible paths towards building machines that learn and think like people. This includes prospects for integrating deep learning with the core cognitive ingredients we identify, inspired in part by recent work fusing neural networks with lower-level building blocks from classic psychology and computer science (attention, working memory, stacks, queues) that have traditionally been seen as incompatible. 

Beyond the specific ingredients in our proposal, we draw a broader distinction between two different computational approaches to intelligence. The statistical \emph{pattern recognition} approach treats prediction as primary, usually in the context of a specific classification, regression, or control task. In this view, learning is about discovering features that have high value states in common -- a shared label in a classification setting or a shared value in a reinforcement learning setting -- across a large, diverse set of training data. The alternative approach treats models of the world as primary, where learning is the process of \emph{model-building}. Cognition is about using these models to understand the world, to explain what we see, to imagine what could have happened that didn't, or what could be true that isn't, and then planning actions to make it so. The difference between pattern recognition and model-building, between prediction and explanation, is central to our view of human intelligence. Just as scientists seek to \emph{explain} nature, not simply predict it, we see human thought as fundamentally a model-building activity. We elaborate this key point with numerous examples below.  We also discuss how pattern recognition, even if it is not the core of intelligence, can nonetheless support model-building, through ``model-free'' algorithms that learn through experience how to make essential inferences more computationally efficient. 

Before proceeding, we provide a few caveats about the goals of this article and a brief overview of the key ideas.

\begin{table}[tbp]
\centering
\caption{\textbf{Glossary}}
\label{glossary}
\begin{tabular}{|p{15cm}|}
\hline
\textbf{Neural network}: A network of simple neuron-like processing units that collectively perform complex computations. Neural networks are often organized into layers, including an input layer that presents the data (e.g, an image), hidden layers that transform the data into intermediate representations, and an output layer that produces a response (e.g., a label or an action). Recurrent connections are also popular when processing sequential data.                                                                                                                                    
\newline 
\textbf{Deep learning}: A neural network with at least one hidden layer (some networks have dozens). Most state-of-the-art deep networks are trained using the backpropagation algorithm to gradually adjust their connection strengths.                                                                                                                                                                                                                                                                                                                                                                   \\ 
\textbf{Backpropagation}: Gradient descent applied to training a deep neural network. The gradient of the objective function (e.g., classification error or log-likelihood) with respect to the model parameters (e.g., connection weights) is used to make a series of small adjustments to the parameters in a direction that improves the objective function.                                                                                                                                                                                                                                           \\ 
\textbf{Convolutional network (convnet)}: A neural network that uses trainable filters instead of (or in addition to) fully-connected layers with independent weights. The same filter is applied at many locations across an image (or across a time series), leading to neural networks that are effectively larger but with local connectivity and fewer free parameters.                                                                                                                                                                                                                               \\ 
\textbf{Model-free and model-based reinforcement learning}: Model-free algorithms directly learn a control policy without explicitly building a model of the environment (reward and state transition distributions). Model-based algorithms learn a model of the environment and use it to select actions by planning.                                                                                                                                                                                                                                                                                                \\ 
\textbf{Deep Q-learning}: A model-free reinforcement learning algorithm used to train deep neural networks on control tasks such as playing Atari games. A network is trained to approximate the optimal action-value function $Q(s,a)$, which is the expected long-term cumulative reward of taking action $a$ in state $s$ and then optimally selecting future actions.                                                                                                                                                                                                                                  \\ 
\textbf{Generative model}: A model that specifies a probability distribution over the data. For instance, in a classification task with examples $X$ and class labels $y$, a generative model specifies the distribution of data given labels $P(X|y)$, as well as a prior on labels $P(y)$, which can be used for sampling new examples or for classification by using Bayes' rule to compute $P(y|X)$. A discriminative model specifies $P(y|X)$ directly, possibly by using a neural network to predict the label for a given data point, and cannot directly be used to sample new examples or to compute other queries regarding the data. We will generally be concerned with directed generative models (such as Bayesian networks or probabilistic programs) which can be given a causal interpretation, although undirected (non-causal) generative models (such as Boltzmann machines) are also possible.\\ 

\textbf{Program induction}: Constructing a program that computes some desired function, where that function is typically specified by training data consisting of example input-output pairs.  In the case of probabilistic programs, which specify candidate generative models for data, an abstract description language is used to define a set of allowable programs and learning is a search for the programs likely to have generated the data.                                                                                                                                                                                                                                                                                                                                                      \\ \hline
\end{tabular}
\end{table}

\subsection{What this article is not}

For nearly as long as there have been neural networks, there have been critiques of neural networks \citep{minsky69,fodor88,pinker88,crick89,Marcus1998,marcus01}. While we are critical of neural networks in this article, our goal is to build on their successes rather than dwell on their shortcomings. We see a role for neural networks in developing more human-like learning machines: They have been applied in compelling ways to many types of machine learning problems, demonstrating the power of gradient-based learning and deep hierarchies of latent variables. Neural networks also have a rich history as computational models of cognition \citep{Rumelhart1986,McClelland1986} -- a history we describe in more detail in the next section. At a more fundamental level, any computational model of learning must ultimately be grounded in the brain's biological neural networks.

We also believe that future generations of neural networks will look very different from the current state-of-the-art. They may be endowed with intuitive physics, theory of mind, causal reasoning, and other capacities we describe in the sections that follow. More structure and inductive biases could be built into the networks or learned from previous experience with related tasks, leading to more human-like patterns of learning and development. Networks may learn to effectively search for and discover new mental models or intuitive theories, and these improved models will, in turn, enable subsequent learning, allowing systems that learn-to-learn -- using previous knowledge to make richer inferences from very small amounts of training data.

It is also important to draw a distinction between AI that purports to emulate or draw inspiration from aspects of human cognition, and AI that does not. This article focuses on the former. The latter is a perfectly reasonable and useful approach to developing AI algorithms -- avoiding cognitive or neural inspiration as well as claims of cognitive or neural plausibility. Indeed, this is how many researchers have proceeded, and this article has little pertinence to work conducted under this research strategy.\footnote{In their influential textbook, \citet{Russell2003} state that ``The quest for `artificial flight' succeeded when the Wright brothers and others stopped imitating birds and started using wind tunnels and learning about aerodynamics.'' (p. 3).} On the other hand, we believe that reverse engineering human intelligence can usefully inform AI and machine learning (and has already done so), especially for the types of domains and tasks that people excel at. Despite recent computational achievements, people are better than machines at solving a range of difficult computational problems, including concept learning, scene understanding, language acquisition, language understanding, speech recognition, etc. Other human cognitive abilities remain difficult to understand computationally, including creativity, common sense, and general purpose reasoning. As long as natural intelligence remains the best example of intelligence, we believe that the project of reverse engineering the human solutions to difficult computational problems will continue to inform and advance AI.

Finally, while we focus on neural network approaches to AI, we do not wish to give the impression that these are the only contributors to recent advances in AI. On the contrary, some of the most exciting recent progress has been in new forms of probabilistic machine learning \citep{ghahramani15}. For example, researchers have developed automated statistical reasoning techniques \citep{lloyd14}, automated techniques for model building and selection \citep{Grosse2012}, and probabilistic programming languages \citep[e.g.,][]{gelman15,Goodman2008,mansinghka14}. We believe that these approaches will play important roles in future AI systems, and they are at least as compatible with the ideas from cognitive science we discuss here, but a full discussion of those connections is beyond the scope of the current article. 

\subsection{Overview of the key ideas}

The central goal of this paper is to propose a set of core ingredients for building more human-like learning and thinking machines. We will elaborate on each of these ingredients and topics in Section \ref{sec:core}, but here we briefly overview the key ideas.

The first set of ingredients focuses on developmental ``start-up software,'' or cognitive capabilities present early in development. There are several reasons for this focus on development. If an ingredient is present early in development, it is certainly active and available well before a child or adult would attempt to learn the types of tasks discussed in this paper. This is true regardless of whether the early-present ingredient is itself learned from experience or innately present. Also, the earlier an ingredient is present, the more likely it is to be foundational to later development and learning.

We focus on two pieces of developmental start-up software \citep[see ][for a review of both]{Wellman1992}. First is \textbf{intuitive physics} (Section \ref{sec:physics}): Infants have primitive object concepts that allow them to track objects over time and allow them to discount physically implausible trajectories. For example, infants know that objects will persist over time and that they are solid and coherent. Equipped with these general principles, people can learn more quickly and make more accurate predictions. While a task may be new, physics still works the same way. A second type of software present in early development is \textbf{intuitive psychology} (Section \ref{sec:psychology}): Infants understand that other people have mental states like goals and beliefs, and this understanding strongly constrains their learning and predictions. A child watching an expert play a new video game can infer that the avatar has agency and is trying to seek reward while avoiding punishment. This inference immediately constrains other inferences, allowing the child to infer what objects are good and what objects are bad. These types of inferences further accelerate the learning of new tasks.

Our second set of ingredients focus on learning. While there are many perspectives on learning, we see \emph{model building} as the hallmark of human-level learning, or explaining observed data through the construction of \textbf{causal} models of the world (Section \ref{sec:causality}). 
Under this perspective, the early-present capacities for intuitive physics and psychology are also causal models of the world. A primary job of learning is to extend and enrich these models, and to build analogous causally structured theories of other domains.

Compared to state-of-the-art algorithms in machine learning, human learning is distinguished by its richness and its efficiency. Children come with the ability and the desire to uncover the underlying causes of sparsely observed events and to use that knowledge to go far beyond the paucity of the data. It might seem paradoxical that people are capable of learning these richly structured models from very limited amounts of experience. We suggest that \textbf{compositionality} and \textbf{learning-to-learn} are ingredients that make this type of rapid model learning possible (Sections \ref{sec:compositionality} and \ref{sec:L2L}, respectively).

A final set of ingredients concerns how the rich models our minds build are put into action, in real time (Section \ref{sec:thinkingfast}). It is remarkable how \emph{fast} we are to perceive and to act. People can comprehend a novel scene in a fraction of a second, and or a novel utterance in little more than the time it takes to say it and hear it. An important motivation for using neural networks in machine vision and speech systems is to respond as quickly as the brain does. Although neural networks are usually aiming at pattern recognition rather than model-building, we will discuss ways in which these ``model-free'' methods can accelerate slow model-based inferences in perception and cognition (Section \ref{sec:inference}). By learning to recognize patterns in these inferences, the outputs of inference can be predicted without having to go through costly intermediate steps. Integrating neural networks that ``learn to do inference" with rich model-building learning mechanisms offers a promising way to explain how human minds can understand the world so well, so quickly.

We will also discuss the integration of model-based and model-free methods in reinforcement learning (Section \ref{sec:RL-2types}), an area that has seen rapid recent progress.  Once a causal model of a task has been learned, humans can use the model to plan action sequences that maximize future reward; when rewards are used as the metric for successs in model-building, this is known as model-based reinforcement learning.  However, planning in complex models is cumbersome and slow, making the speed-accuracy trade-off unfavorable for real-time control. By contrast, model-free reinforcement learning algorithms, such as current instantiations of deep reinforcement learning, support fast control but at the cost of inflexibility and possibly accuracy. We will review evidence that humans combine model-based and model-free learning algorithms both competitively and cooperatively, and that these interactions are supervised by metacognitive processes. The sophistication of human-like reinforcement learning has yet to be realized in AI systems, but this is an area where crosstalk between cognitive and engineering approaches is especially promising.

\section{Cognitive and neural inspiration in artificial intelligence}
The questions of whether and how AI should relate to human cognitive psychology are older than the terms `artificial intelligence' and `cognitive psychology.' Alan Turing suspected that it is easier to build and educate a child-machine than try to fully capture adult human cognition \citep{Turing1950a}. Turing pictured the child's mind as a notebook with ``rather little mechanism and lots of blank sheets,'' and the mind of a child-machine as filling in the notebook by responding to rewards and punishments, similar to reinforcement learning. This view on representation and learning echoes behaviorism, a dominant psychological tradition in Turing's time. It also echoes the strong empiricism of modern connectionist models, the idea that we can learn almost everything we know from the statistical patterns of sensory inputs.

Cognitive science repudiated the over-simplified behaviorist view and came to play a central role in early AI research \citep{boden06}. \citet{newell61} developed their ``General Problem Solver'' as both an AI algorithm and a model of human problem solving, which they subsequently tested experimentally \citep{newell72}. AI pioneers in other areas of research explicitly referenced human cognition, and even published papers in cognitive psychology journals \citep[e.g.,][]{bobrow77,hayes79,winograd72}. For example, \citet{schank72}, writing in the journal \emph{Cognitive Psychology}, declared that
\begin{quotation}
\noindent \textit{We hope to be able to build a program that can learn, as a child does, how to do what we have described in this paper instead of being spoon-fed the tremendous information necessary}.
\end{quotation}
A similar sentiment was expressed by \citet{minsky74}:
\begin{quotation}
\noindent \textit{I draw no boundary between a theory of human thinking and a scheme for making an intelligent machine; no purpose would be served by separating these today since neither domain has theories good enough to explain---or to produce---enough mental capacity.}
\end{quotation}
Much of this research assumed that human knowledge representation is symbolic and that reasoning, language, planning and vision could be understood in terms of symbolic operations. Parallel to these developments, a radically different approach was being explored, based on neuron-like ``sub-symbolic'' computations \citep[e.g.,][]{fukushima80,grossberg76,rosenblatt58}. The representations and algorithms used by this approach were more directly inspired by neuroscience than by cognitive psychology, although ultimately it would flower into an influential school of thought about the nature of cognition---\emph{parallel distributed processing} (PDP) \citep{Rumelhart1986,McClelland1986}. As its name suggests, PDP emphasizes parallel computation by combining simple units to collectively implement sophisticated computations. The knowledge learned by these neural networks is thus distributed across the collection of units rather than localized as in most symbolic data structures. The resurgence of recent interest in neural networks, more commonly referred to as ``deep learning,'' share the same representational commitments and often even the same learning algorithms as the earlier PDP models. ``Deep'' refers to the fact that more powerful models can be built by composing many layers of representation \citep[see][for recent reviews]{Lecun2015,schmidhuber15}, still very much in the PDP style while utilizing recent advances in hardware and computing capabilities, as well as massive datasets, to learn deeper models.

It is also important to clarify that the PDP perspective is compatible with ``model building'' in addition to ``pattern recognition.'' Some of the original work done under the banner of PDP \citep{Rumelhart1986} is closer to model building than pattern recognition, whereas the recent large-scale discriminative deep learning systems more purely exemplify pattern recognition \citep[see ][ for a related discussion]{bottou2014}. But, as discussed, there is also a question of the nature of the learned representations within the model -- their form, compositionality, and transferability -- and the developmental start-up software that was used to get there. We focus on these issues in this paper.

Neural network models and the PDP approach offer a view of the mind (and intelligence more broadly) that is sub-symbolic and often populated with minimal constraints and inductive biases to guide learning. Proponents of this approach maintain that many classic types of structured knowledge, such as graphs, grammars, rules, objects, structural descriptions, programs, etc. can be useful yet misleading metaphors for characterizing thought. These structures are more epiphenomenal than real, emergent properties of more fundamental sub-symbolic cognitive processes \citep{McClelland2010}. Compared to other paradigms for studying cognition, this position on the nature of representation is often accompanied by a relatively ``blank slate'' vision of initial knowledge and representation, much like Turing's blank notebook.

When attempting to understand a particular cognitive ability or phenomenon within this paradigm, a common scientific strategy is to train a relatively generic neural network to perform the task, adding additional ingredients only when necessary. This approach has shown that neural networks can behave as if they learned explicitly structured knowledge, such as a rule for producing the past tense of words \citep{Rumelhart1986b}, rules for solving simple balance-beam physics problems \citep{mcclelland1988parallel}, or a tree to represent types of living things (plants and animals) and their distribution of properties \citep{Rogers2004}. Training large-scale relatively generic networks is also the best current approach for object recognition \citep{Krizhevsky2012,Russakovsky2014,Szegedy2014,He2015}, where the high-level feature representations of these convolutional nets have also been used to predict patterns of neural response in human and macaque IT cortex \citep{Yamins2014,Khaligh-Razavi2014,Kriegeskorte2015} as well as human typicality ratings \citep{Lake2014b} and similarity ratings \citep{Peterson2016} for images of common objects. Moreover, researchers have trained generic networks to perform structured and even strategic tasks, such as the recent work on using a Deep Q-learning Network (DQN) to play simple video games \citep{Mnih2015}. If neural networks have such broad application in machine vision, language, and control, and if they can be trained to emulate the rule-like and structured behaviors that characterize cognition, do we need more to develop truly human-like learning and thinking machines? How far can relatively generic neural networks bring us towards this goal?

\section{Challenges for building more human-like machines}

While cognitive science has not yet converged on a single account of the mind or intelligence, the claim that a mind is a collection of general purpose neural networks with few initial constraints is rather extreme in contemporary cognitive science. A different picture has emerged that highlights the importance of early inductive biases, including core concepts such as number, space, agency and objects, as well as powerful learning algorithms that rely on prior knowledge to extract knowledge from small amounts of training data. This knowledge is often richly organized and theory-like in structure, capable of the graded inferences and productive capacities characteristic of human thought.

Here we present two challenge problems for machine learning and AI: learning simple visual concepts \citep{LakeScience2015} and learning to play the Atari game Frostbite \citep{Mnih2015}. We also use the problems as running examples to illustrate the importance of core cognitive ingredients in the sections that follow.

\subsection{The Characters Challenge} \label{characters_section}
The first challenge concerns handwritten character recognition, a classic problem for comparing different types of machine learning algorithms. \citet{Hofstadter1985} argued that the problem of recognizing characters in all the ways people do -- both handwritten and printed -- contains most if not all of the fundamental challenges of AI. Whether or not this statement is right, it highlights the surprising complexity that underlies even ``simple'' human-level concepts like letters. More practically, handwritten character recognition is a real problem that children and adults must learn to solve, with practical applications ranging from reading envelope addresses or checks in an ATM machine. Handwritten character recognition is also simpler than more general forms of object recognition -- the object of interest is two-dimensional, separated from the background, and usually unoccluded. Compared to how people learn and see other types of objects, it seems possible, in the near term, to build algorithms that can see most of the structure in characters that people can see.

The standard benchmark is the MNIST data set for digit recognition, which involves classifying images of digits into the categories `0'-`9' \citep{LeCunBottou1998}. The training set provides 6,000 images per class for a total of 60,000 training images. With a large amount of training data available, many algorithms achieve respectable performance, including K-nearest neighbors (5\% test error), support vector machines (about 1\% test error), and convolutional neural networks \citep[below 1\% test error;][]{LeCunBottou1998}. The best results achieved using deep convolutional nets are very close to human-level performance at an error rate of 0.2\% \citep{Ciresan2012}. Similarly, recent results applying convolutional nets to the far more challenging ImageNet object recognition benchmark have shown that human-level performance is within reach on that data set as well \citep{Russakovsky2014}.

\begin{figure}
    \centering    
    \includegraphics[width=5.5in]{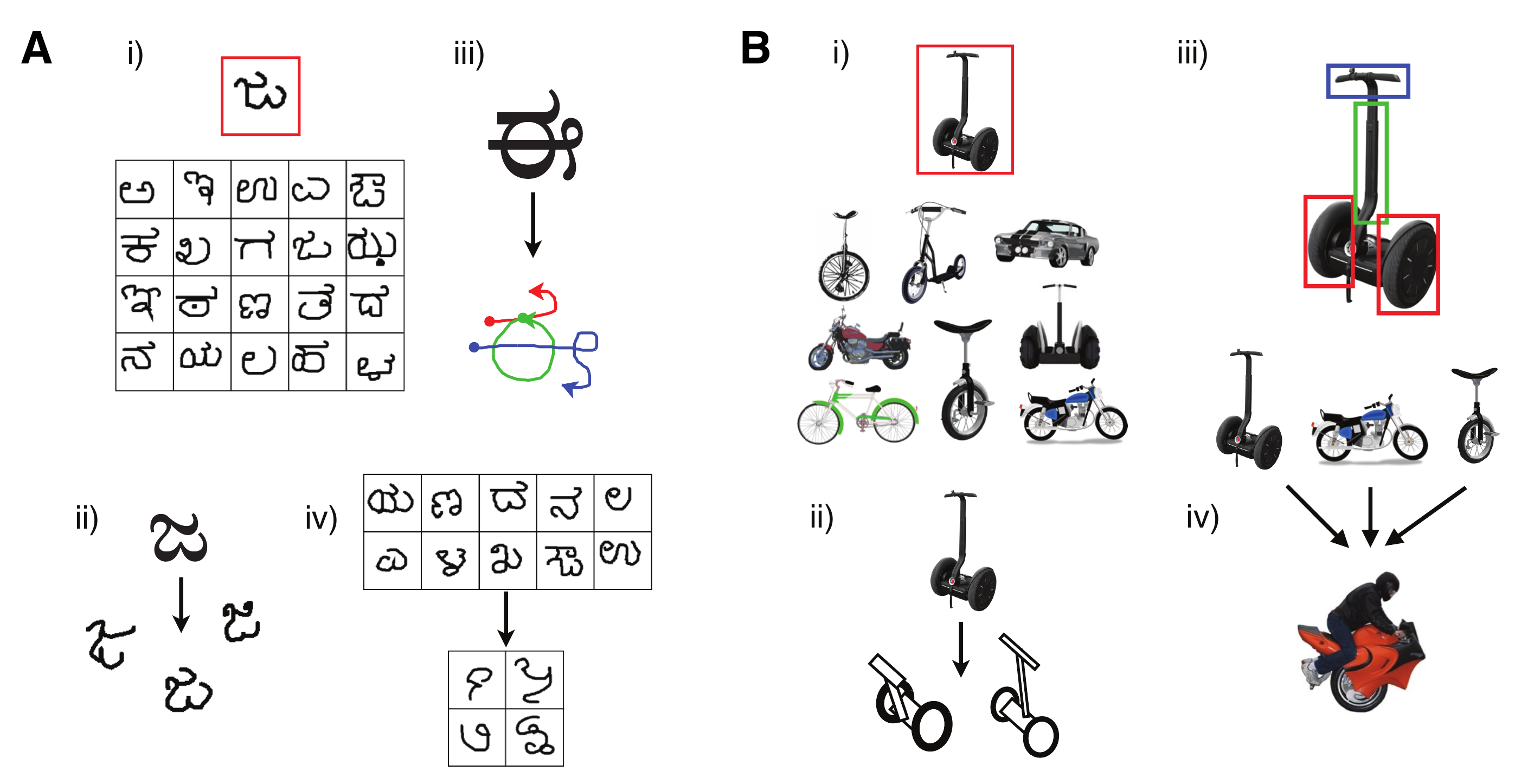}
    \caption{The characters challenge: human-level learning of a novel handwritten characters (A), with the same abilities also illustrated for a novel two-wheeled vehicle (B). A single example of a new visual concept (red box) can be enough information to support the (i) classification of new examples, (ii) generation of new examples, (iii) parsing an object into parts and relations, and (iv) generation of new concepts from related concepts. Adapted from \citet{LakeScience2015}.}
    \label{fig_characters_challenge}
\end{figure}

While humans and neural networks may perform equally well on the MNIST digit recognition task and other large-scale image classification tasks, it does not mean that they learn and think in the same way. There are at least two important differences: people learn from fewer examples and they learn richer representations, a comparison true for both learning handwritten characters as well as learning more general classes of objects (Figure \ref{fig_characters_challenge}). People can learn to recognize a new handwritten character from a single example (Figure \ref{fig_characters_challenge}A-i), allowing them to discriminate between novel instances drawn by other people and similar looking non-instances \citep{MillerMatsakis2000,LakeScience2015}. Moreover, people learn more than how to do pattern recognition: they learn a concept -- that is, a model of the class that allows their acquired knowledge to be flexibly applied in new ways. In addition to recognizing new examples, people can also generate new examples (Figure \ref{fig_characters_challenge}A-ii), parse a character into its most important parts and relations (Figure  \ref{fig_characters_challenge}A-iii; \citet{Lake2012}), and generate new characters given a small set of related characters (Figure  \ref{fig_characters_challenge}A-iv). These additional abilities come for free along with the acquisition of the underlying concept.

Even for these simple visual concepts, people are still better and more sophisticated learners than the best algorithms for character recognition. People learn a lot more from a lot less, and capturing these human-level learning abilities in machines is the \emph{Characters Challenge}. We recently reported progress on this challenge using probabilistic program induction \citep{LakeScience2015}, yet aspects of the full human cognitive ability remain out of reach. While both people and model represent characters as a sequence of pen strokes and relations, people have a far richer repertoire of structural relations between strokes. Furthermore, people can efficiently integrate across multiple examples of a character to infer which have optional elements, such as the horizontal cross-bar in `7's, combining different variants of the same character into a single coherent representation. Additional progress may come by combining deep learning and probabilistic program induction to tackle even richer versions of the Characters Challenge.

\subsection{The Frostbite Challenge} \label{frostbite_section}

\begin{figure}
\centering\includegraphics[width=6.5in]{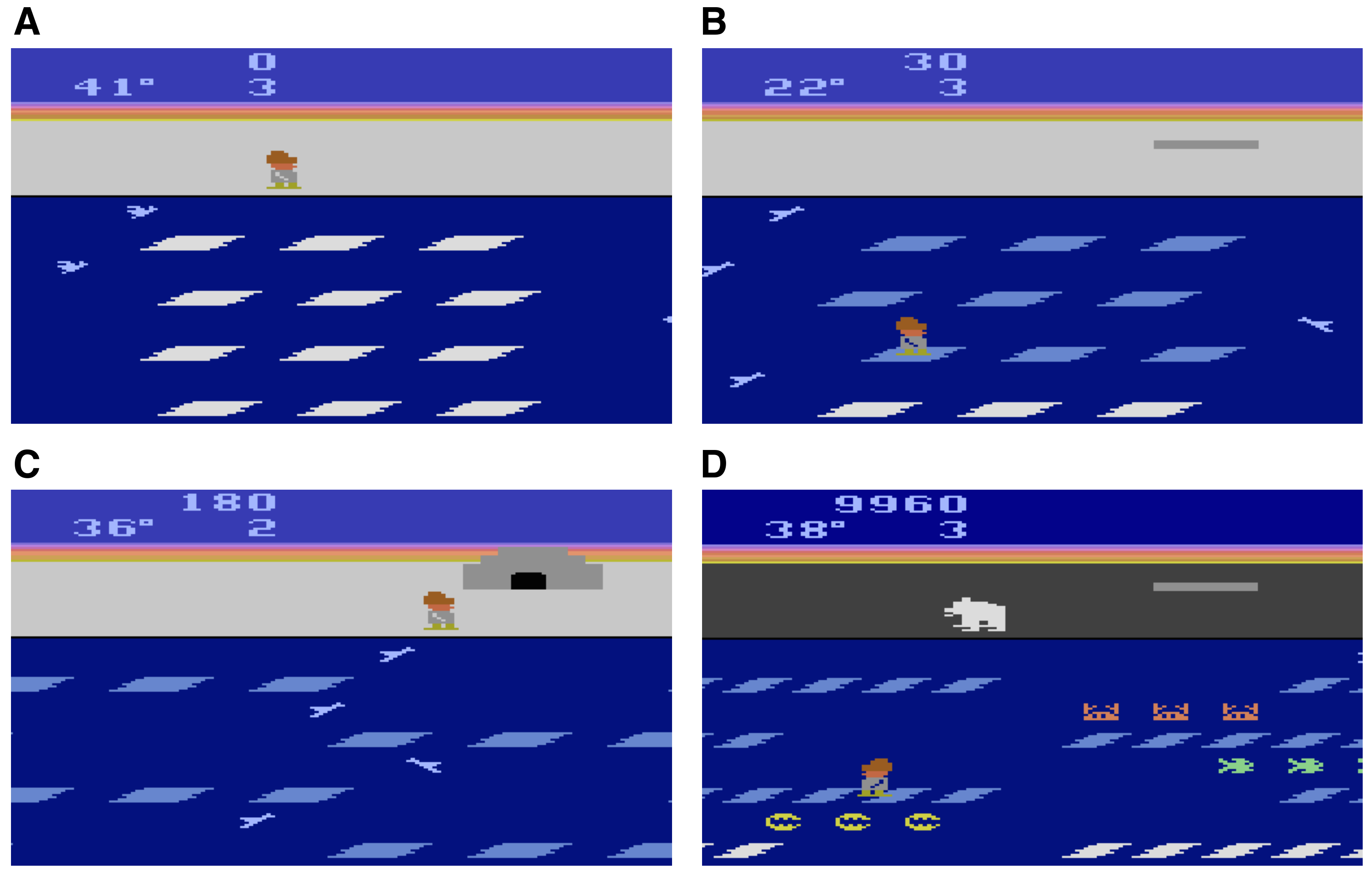}
\caption{Screenshots of Frostbite, a 1983 video game designed for the Atari game console. A) The start of a level in Frostbite. The agent must construct an igloo by hopping between ice floes and avoiding obstacles such as birds. The floes are in constant motion (either left or right), making multi-step planning essential to success. B) The agent receives pieces of the igloo (top right) by jumping on the active ice floes (white), which then deactivates them (blue). C) At the end of a level, the agent must safely reach the completed igloo. D) Later levels include additional rewards (fish) and deadly obstacles (crabs, clams, and bears).}
\label{frostbite}
\end{figure}

The second challenge concerns the Atari game Frostbite (Figure \ref{frostbite}), which was one of the control problems tackled by the DQN of \citet{Mnih2015}. The DQN was a significant advance in reinforcement learning, showing that a single algorithm can learn to play a wide variety of complex tasks. The network was trained to play 49 classic Atari games, proposed as a test domain for reinforcement learning \citep{Bellemare2015}, impressively achieving human-level performance or above on 29 of the games. It did, however, have particular trouble with Frostbite and other games that required temporally extended planning strategies.

In Frostbite, players control an agent (Frostbite Bailey) tasked with constructing an igloo within a time limit. The igloo is built piece-by-piece as the agent jumps on ice floes in water (Figure \ref{frostbite}A-C). The challenge is that the ice floes are in constant motion (moving either left or right), and ice floes only contribute to the construction of the igloo if they are visited in an active state (white rather than blue). The agent may also earn extra points by gathering fish while avoiding a number of fatal hazards (falling in the water, snow geese, polar bears, etc.). Success in this game requires a temporally extended plan to ensure the agent can accomplish a sub-goal (such as reaching an ice floe) and then safely proceed to the next sub-goal. Ultimately, once all of the pieces of the igloo are in place, the agent must proceed to the igloo and thus complete the level before time expires (Figure \ref{frostbite}C).

The DQN learns to play Frostbite and other Atari games by combining a powerful pattern recognizer (a deep convolutional neural network) and a simple model-free reinforcement learning algorithm \citep[Q-learning;][]{watkins92}. These components allow the network to map sensory inputs (frames of pixels) onto a policy over a small set of actions, and both the mapping and the policy are trained to optimize long-term cumulative reward (the game score). The network embodies the strongly empiricist approach characteristic of most connectionist models: very little is built into the network apart from the assumptions about image structure inherent in convolutional networks, so the network has to essentially learn a visual and conceptual system from scratch for each new game. In \citet{Mnih2015}, the network architecture and hyper-parameters were fixed, but the network was trained anew for each game, meaning the visual system and the policy are highly specialized for the games it was trained on. More recent work has shown how these game-specific networks can share visual features \citep{Rusu2016} or be used to train a multi-task network \citep{Parisotto2015}, achieving modest benefits of transfer when learning to play new games.

Although it is interesting that the DQN learns to play games at human-level performance while assuming very little prior knowledge, the DQN may be learning to play Frostbite and other games in a very different way than people do. One way to examine the differences is by considering the amount of experience required for learning. In \citet{Mnih2015}, the DQN was compared with a professional gamer who received approximately two hours of practice on each of the 49 Atari games (although he or she likely had prior experience with some of the games). The DQN was trained on 200 million frames from each of the games, which equates to approximately 924 hours of game time (about 38 days), or almost 500 times as much experience as the human received.\footnote{The time required to train the DQN (compute time) is not the same as the game (experience) time. Compute time can be longer.} Additionally, the DQN incorporates experience replay, where each of these frames is replayed approximately 8 more times on average over the course of learning.

\begin{figure}
\centering\includegraphics[width=4in]{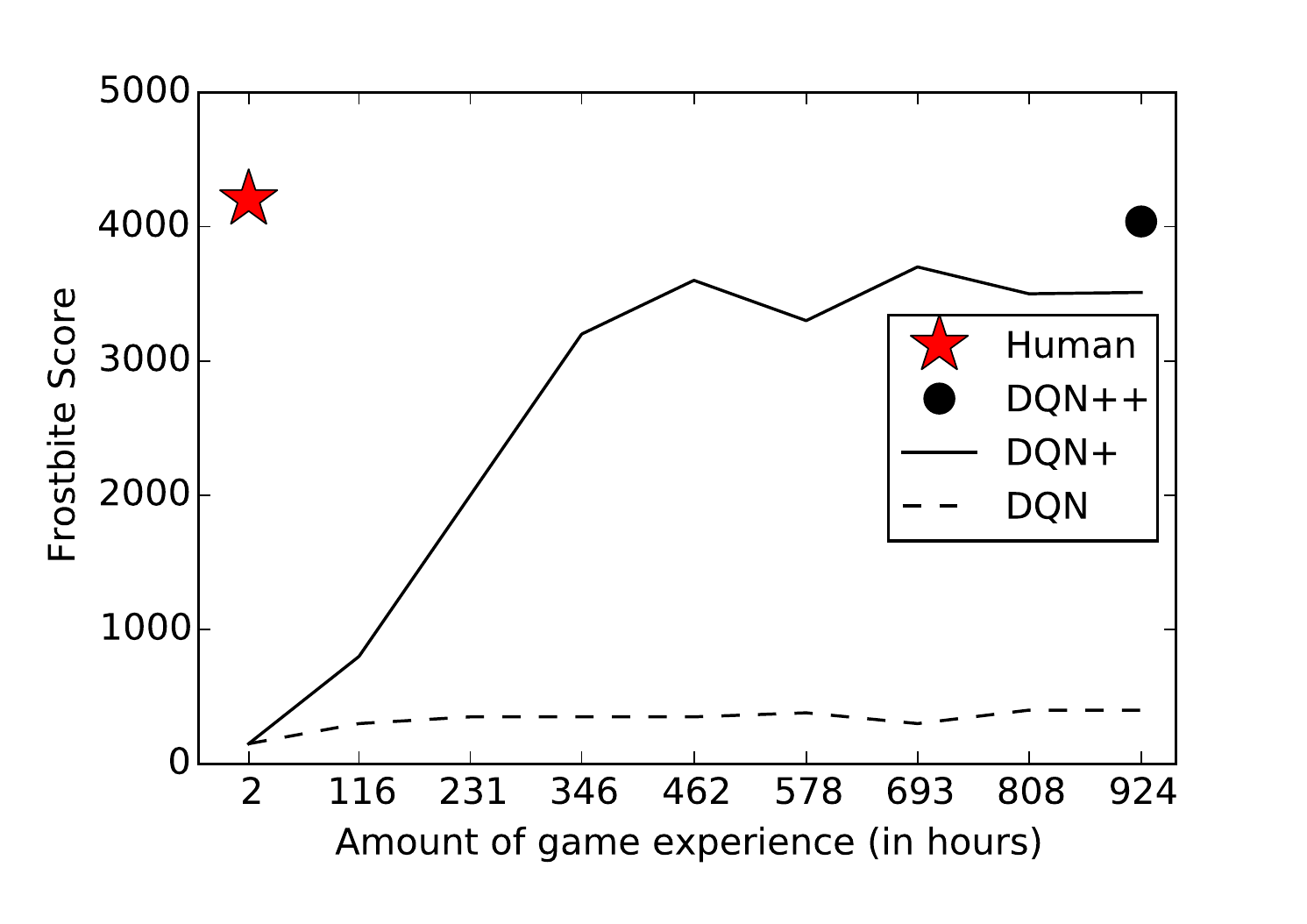}
\caption{Comparing learning speed for people versus Deep Q-Networks (DQNs). Test performance on the Atari 2600 game ``Frostbite'' is plotted as a function of game experience (in hours at a frame rate of 60 fps), which does not include additional experience replay. Learning curves (if available) and scores are shown from different networks: DQN \citep{Mnih2015}, DQN+ \citep{Schaul2015}, and DQN++ \citep{Wang2016}. Random play achieves a score of 66.4. The ``human starts'' performance measure is used \citep{VanHasselt2015}.}
\label{frostbite_learning}
\end{figure}

With the full 924 hours of unique experience and additional replay, the DQN achieved less than 10\% of human-level performance during a controlled test session (see DQN in Fig. \ref{frostbite_learning}). More recent variants of the DQN have demonstrated superior performance \citep{Stadie2015,VanHasselt2015,Schaul2015,Wang2016}, reaching 83\% of the professional gamer's score by incorporating smarter experience replay \citep{Schaul2015} and 96\% by using smarter replay and more efficient parameter sharing \citep{Wang2016} (see DQN+ and DQN++ in Fig. \ref{frostbite_learning}).\footnote{The reported scores use the ``human starts'' measure of test performance, designed to prevent networks from just memorizing long sequences of successful actions from a single starting point. Both faster learning \citep{Blundell2016} and higher scores \citep{Wang2016} have been reported using other metrics, but it is unclear how well the networks are generalizing with these alternative metrics.} But they requires a lot of experience to reach this level: the learning curve provided in \citet{Schaul2015} shows performance is around 46\% after 231 hours, 19\% after 116 hours, and below 3.5\% after just 2 hours (which is close to random play, approximately 1.5\%). The differences between the human and machine learning curves suggest that they may be learning different kinds of knowledge, using different learning mechanisms, or both.

The contrast becomes even more dramatic if we look at the very earliest stages of learning. While both the original DQN and these more recent variants require multiple hours of experience to perform reliably better than random play, even non-professional humans can grasp the basics of the game after just a few minutes of play. We speculate that people do this by inferring a general schema to describe the goals of the game and the object types and their interactions, using the kinds of intuitive theories, model-building abilities and model-based planning mechanisms we describe below.  While novice players may make some mistakes, such as inferring that fish are harmful rather than helpful, they can learn to play better than chance within a few minutes.  If humans are able to first watch an expert playing for a few minutes, they can learn even faster.  In informal experiments with two of the authors playing Frostbite on a Javascript emulator (http://www.virtualatari.org/soft.php?soft=Frostbite), after watching videos of expert play on YouTube for just two minutes, we found that we were able to reach scores comparable to or better than the human expert reported in \citet{Mnih2015} after at most 15-20 minutes of total practice.\footnote{More precisely, the human expert in \citet{Mnih2015} scored an average of 4335 points across 30 game sessions of up to five minutes of play. In individual sessions lasting no longer than five minutes, author TDU obtained scores of 3520 points after approximately 5 minutes of gameplay, 3510 points after 10 minutes, and 7810 points after 15 minutes.  Author JBT obtained 4060 after approximately 5 minutes of gameplay, 4920 after 10-15 minutes, and 6710 after no more than 20 minutes.  TDU and JBT each watched approximately two minutes of expert play on YouTube (e.g., https://www.youtube.com/watch?v=ZpUFztf9Fjc, but there are many similar examples that can be found in a YouTube search).}  

There are other behavioral signatures that suggest fundamental differences in representation and learning between people and the DQN. For instance, the game of Frostbite provides incremental rewards for reaching each active ice floe, providing the DQN with the relevant sub-goals for completing the larger task of building an igloo. Without these sub-goals, the DQN would have to take random actions until it accidentally builds an igloo and is rewarded for completing the entire level. In contrast, people likely do not rely on incremental scoring in the same way when figuring out how to play a new game. In Frostbite, it is possible to figure out the higher-level goal of building an igloo without incremental feedback; similarly, sparse feedback is a source of difficulty in other Atari 2600 games such as Montezuma's Revenge where people substantially outperform current DQN approaches.

The learned DQN network is also rather inflexible to changes in its inputs and goals: changing the color or appearance of objects or changing the goals of the network would have devastating consequences on performance if the network is not retrained. While any specific model is necessarily simplified and should not be held to the standard of general human intelligence, the contrast between DQN and human flexibility is striking nonetheless. For example, imagine you are tasked with playing Frostbite with any one of these new goals:
\begin{itemize}
\item Get the lowest possible score.
\item Get closest to 100, or 300, or 1000, or 3000, or any level, without going over. 
\item Beat your friend, who's playing next to you, but just barely, not by too much, so as not to embarrass them. 
\item Go as long as you can without dying. 
\item Die as quickly as you can. 
\item Pass each level at the last possible minute, right before the temperature timer hits zero and you die (i.e., come as close as you can to dying from frostbite without actually dying).
\item Get to the furthest unexplored level without regard for your score.
\item See if you can discover secret Easter eggs.
\item Get as many fish as you can.
\item Touch all the individual ice floes on screen once and only once.
\item Teach your friend how to play as efficiently as possible.
\end{itemize}
This range of goals highlights an essential component of human intelligence: people can learn models and use them for arbitrary new tasks and goals.  While neural networks can learn multiple mappings or tasks with the same set of stimuli -- adapting their outputs depending on a specified goal -- these models require substantial training or reconfiguration to add new tasks \citep[e.g.,][]{Rougier2005,Eliasmith2012,Collins2013}. In contrast, people require little or no retraining or reconfiguration, adding new tasks and goals to their repertoire with relative ease.

The Frostbite example is a particularly telling contrast when compared with human play.  Even the best deep networks learn gradually over many thousands of game episodes, take a long time to reach good performance and are locked into particular input and goal patterns. Humans, after playing just a small number of games over a span of minutes, can understand the game and its goals well enough to perform better than deep networks do after almost a thousand hours of experience. Even more impressively, people understand enough to invent or accept new goals, generalize over changes to the input, and explain the game to others. Why are people different? What core ingredients of human intelligence might the DQN and other modern machine learning methods be missing?

One might object that both the Frostbite and Characters challenges draw an unfair comparison between the speed of human learning and neural network learning. We discuss this objection in detail in Section \ref{sec:objections}, but we feel it is important to anticipate here as well. To paraphrase one reviewer of an earlier draft of this article, ``It is not that DQN and people are solving the same task differently. They may be better seen as solving different tasks. Human learners -- unlike DQN and many other deep learning systems -- approach new problems armed with extensive prior experience. The human is encountering one in a years-long string of problems, with rich overlapping structure. Humans as a result often have important domain-specific knowledge for these tasks, even before they `begin.' The DQN is starting completely from scratch.'' We agree, and indeed this is another way of putting our point here. Human learners fundamentally take on different learning tasks than today's neural networks, and if we want to build machines that learn and think like people, our machines need to confront the kinds of tasks that human learners do, not shy away from them. People never start completely from scratch, or even close to ``from scratch,'' and that is the secret to their success.  The challenge of building models of human learning and thinking then becomes: How do we bring to bear rich prior knowledge to learn new tasks and solve new problems so quickly? What form does that prior knowledge take, and how is it constructed, from some combination of inbuilt capacities and previous experience? The core ingredients we propose in the next section offer one route to meeting this challenge.

\section{Core ingredients of human intelligence} \label{sec:core}
In the Introduction, we laid out what we see as core ingredients of intelligence. Here we consider the ingredients in detail and contrast them with the current state of neural network modeling. While these are hardly the only ingredients needed for human-like learning and thought (see our discussion of language in Section \ref{sec:objections}), they are key building blocks which are not present in most current learning-based AI systems -- certainly not all present together -- and for which additional attention may prove especially fruitful. We believe that integrating them will produce significantly more powerful and more human-like learning and thinking abilities than we currently see in AI systems.

Before considering each ingredient in detail, it is important to clarify that by ``core ingredient'' we do not necessarily mean an ingredient that is innately specified by genetics or must be ``built in'' to any learning algorithm. We intend our discussion to be agnostic with regards to the origins of the key ingredients. By the time a child or an adult is picking up a new character or learning how to play Frostbite, they are armed with extensive real world experience that deep learning systems do not benefit from -- experience that would be hard to emulate in any general sense. Certainly, the core ingredients are enriched by this experience, and some may even be a product of the experience itself. Whether learned, built in, or enriched, the key claim is that these ingredients play an active and important role in producing human-like learning and thought, in ways contemporary machine learning has yet to capture.

\subsection{Developmental start-up software}
\label{sec_dev_start}

Early in development, humans have a foundational understanding of several core domains \citep{Spelke2007,spelke2003core}. These domains include number (numerical and set operations), space (geometry and navigation), physics (inanimate objects and mechanics) and psychology (agents and groups). These core domains cleave cognition at its conceptual joints, and each domain is organized by a set of entities and abstract principles relating the entities. The underlying cognitive representations can be understood as ``intuitive theories,'' with a causal structure resembling a scientific theory \citep{Carey2004,Carey2009,Gopnik1999,Gopnik2004,Schulz2012,Wellman1992,Wellman1998,Gweon2010}. The ``child as scientist'' proposal further views the process of learning itself as also scientist-like, with recent experiments showing that children seek out new data to distinguish between hypotheses, isolate variables, test causal hypotheses, make use of the data-generating process in drawing conclusions, and learn selectively from others \citep{Gweon2010,Cook2011,Schulz2007a,Stahl2015,Tsividis2013}. We will address the nature of learning mechanisms in Section \ref{sec:learning}.

Each core domain has been the target of a great deal of study and analysis, and together the domains are thought to be shared cross-culturally and partly with non-human animals. All of these domains may be important augmentations to current machine learning, though below we focus in particular on the early understanding of objects and agents.

\subsubsection{Intuitive physics} \label{sec:physics}

Young children have rich knowledge of intuitive physics. Whether learned or innate, important physical concepts are present at ages far earlier than when a child or adult learns to play Frostbite, suggesting these resources may be used for solving this and many everyday physics-related tasks.

At the age of 2 months and possibly earlier, human infants expect inanimate objects to follow principles of persistence, continuity, cohesion and solidity. Young infants believe objects should move along smooth paths, not wink in and out of existence, not inter-penetrate and not act at a distance \citep{Spelke1995,Spelke1990}. These expectations guide object segmentation in early infancy, emerging before appearance-based cues such as color, texture, and perceptual goodness \citep{Spelke1990}.

These expectations also go on to guide later learning. At around 6 months, infants have already developed different expectations for rigid bodies, soft bodies and liquids \citep{rips2015divisions}. Liquids, for example, are expected to go through barriers, while solid objects cannot \citep{Hespos2009}. By their first birthday, infants have gone through several transitions of comprehending basic physical concepts such as inertia, support, containment and collisions \citep{Hespos2008,Baillargeon2004,baillargeon2009account}.

There is no single agreed-upon computational account of these early physical principles and concepts, and previous suggestions have ranged from decision trees \citep{baillargeon2009account}, to cues, to lists of rules \citep{siegler1998developmental}. A promising recent approach sees intuitive physical reasoning as similar to inference over a physics software engine, the kind of simulators that power modern-day animations and games \citep{Sanborn2013,Battaglia2013,Bates2015,Gerstenberg2015}. According to this hypothesis, people reconstruct a perceptual scene using internal representations of the objects and their physically relevant properties (such as mass, elasticity, and surface friction), and forces acting on objects (such as gravity, friction, or collision impulses). Relative to physical ground truth, the intuitive physical state representation is approximate and probabilistic, and oversimplified and incomplete in many ways.  Still, it is rich enough to support mental simulations that can predict how objects will move in the immediate future, either on their own or in responses to forces we might apply. 

\begin{figure}
\centering\includegraphics[width=6.in]{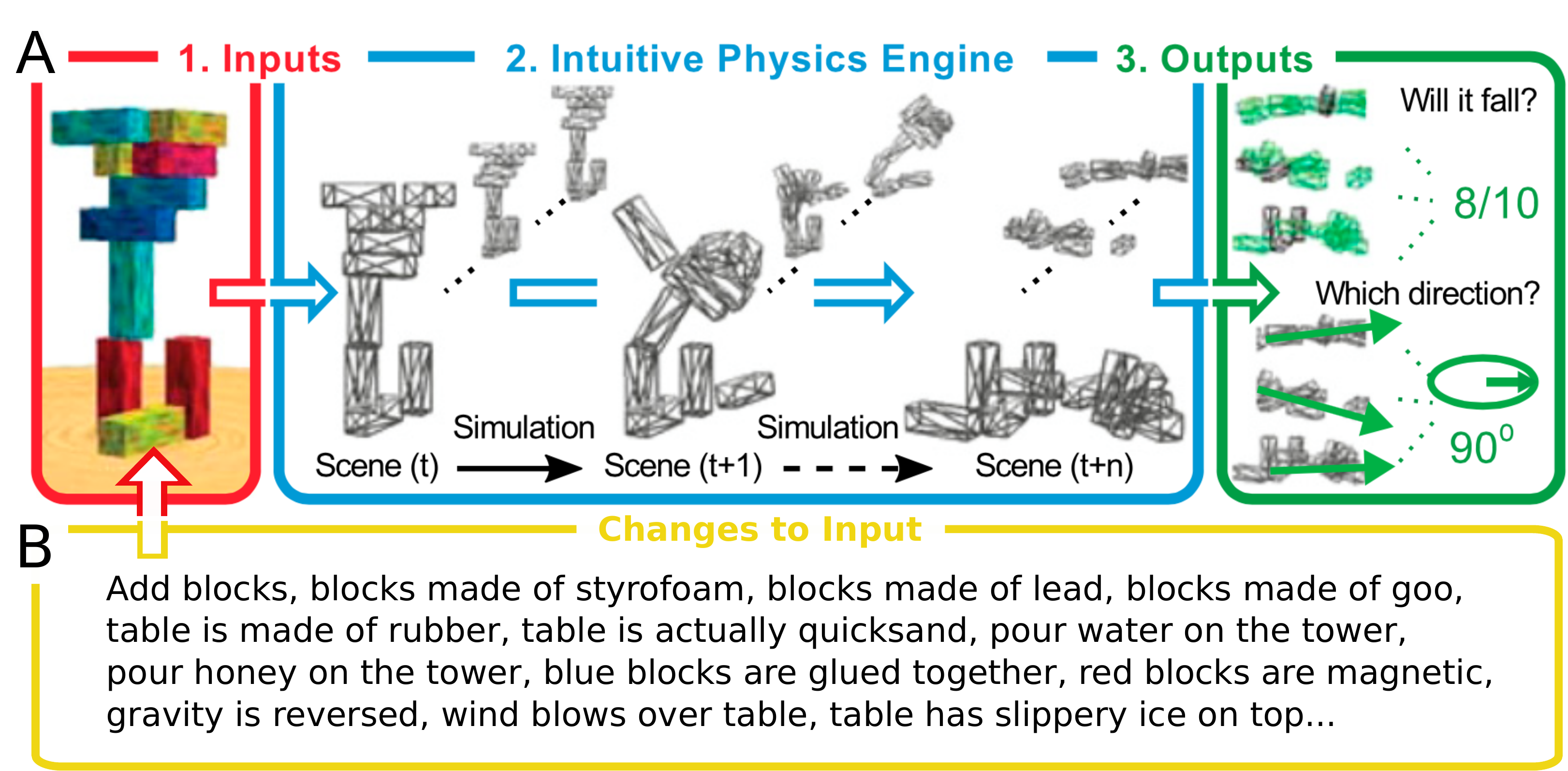}
\caption{The intuitive physics-engine approach to scene understanding, illustrated through tower stability.  (A) The engine takes in inputs through perception, language, memory and other faculties. It then constructs a physical scene with objects, physical properties and forces, simulates the scene's development over time and hands the output to other reasoning systems. (B) Many possible `tweaks' to the input can result in much different scenes, requiring the potential discovery, training and evaluation of new features for each tweak. Adapted from \citet{Battaglia2013}.}
\label{fig:physics}
\end{figure}

This ``intuitive physics engine'' approach enables flexible adaptation to a wide range of everyday scenarios and judgments in a way that goes beyond perceptual cues.  For example (Figure \ref{fig:physics}), a physics-engine reconstruction of a tower of wooden blocks from the game Jenga can be used to predict whether (and how) a tower will fall, finding close quantitative fits to how adults make these predictions \citep{Battaglia2013} as well as simpler kinds of physical predictions that have been studied in infants \citep{Teglas2011}. Simulation-based models can also capture how people make hypothetical or counterfactual predictions: What would happen if certain blocks are taken away, more blocks are added, or the table supporting the tower is jostled?  What if certain blocks were glued together, or attached to the table surface?  What if the blocks were made of different materials (Styrofoam, lead, ice)?  What if the blocks of one color were much heavier than other colors? Each of these physical judgments may require new features or new training for a pattern recognition account to work at the same level as the model-based simulator. 

What are the prospects for embedding or acquiring this kind of intuitive physics in deep learning systems?  Connectionist models in psychology have previously been applied to physical reasoning tasks such as balance-beam rules \citep{mcclelland1988parallel,shultz2003computational} or rules relating distance, velocity, and time in motion \citep{buckingham2000developmental}, but these networks do not attempt to work with complex scenes as input or a wide range of scenarios and judgments as in Figure \ref{fig:physics}. A recent paper from Facebook AI researchers \citep{Lerer2016} represents an exciting step in this direction. \citet{Lerer2016} trained a deep convolutional network-based system (PhysNet) to predict the stability of block towers from simulated images similar to those in Figure \ref{fig:physics}A but with much simpler configurations of two, three or four cubical blocks stacked vertically. Impressively, PhysNet generalized to simple real images of block towers, matching human performance on these images, meanwhile exceeding human performance on synthetic images. Human and PhysNet confidence were also correlated across towers, although not as strongly as for the approximate probabilistic simulation models and experiments of \citet{Battaglia2013}. One limitation is that PhysNet currently requires extensive training -- between 100,000 and 200,000 scenes -- to learn judgments for just a single task (will the tower fall?) on a narrow range of scenes (towers with two to four cubes).  It has been shown to generalize, but also only in limited ways (e.g., from towers of two and three cubes to towers of four cubes). In contrast, people require far less experience to perform any particular task, and can generalize to many novel judgments and complex scenes with no new training required (although they receive large amounts of physics experience through interacting with the world more generally). Could deep learning systems such as PhysNet capture this flexibility, without explicitly simulating the causal interactions between objects in three dimensions?  We are not sure, but we hope this is a challenge they will take on.

Alternatively, instead of trying to make predictions without simulating physics, could neural networks be trained to emulate a general-purpose physics simulator, given the right type and quantity of training data, such as the raw input experienced by a child?  This is an active and intriguing area of research, but it too faces significant challenges.  For networks trained on object classification, deeper layers often become sensitive to successively higher-level features, from edges to textures to shape-parts to full objects \citep{Zeiler2012,Yosinski2014}. For deep networks trained on physics-related data, it remains to be seen whether higher layers will encode objects, general physical properties, forces and approximately Newtonian dynamics.  A generic network trained on dynamic pixel data might learn an implicit representation of these concepts, but would it generalize broadly beyond training contexts as people's more explicit physical concepts do?  Consider for example a network that learns to predict the trajectories of several balls bouncing in a box \citep{Michalski2014}. If this network has actually learned something like Newtonian mechanics, then it should be able to generalize to interestingly different scenarios -- at a minimum different numbers of differently shaped objects, bouncing in boxes of different shapes and sizes and orientations with respect to gravity, not to mention more severe generalization tests such as all of the tower tasks discussed above, which also fall under the Newtonian domain. Neural network researchers have yet to take on this challenge, but we hope they will.  Whether such models can be learned with the kind (and quantity) of data available to human infants is not clear, as we discuss further in Section \ref{sec:objections}. 

It may be difficult to integrate object and physics-based primitives into deep neural networks, but the payoff in terms of learning speed and performance could be great for many tasks. Consider the case of learning to play Frostbite.  Although it can be difficult to discern exactly how a network learns to solve a particular task, the DQN probably does not parse a Frostbite screenshot in terms of stable objects or sprites moving according to the rules of intuitive physics (Figure \ref{frostbite}). But incorporating a physics-engine-based representation could help DQNs learn to play games such as Frostbite in a faster and more general way, whether the physics knowledge is captured implicitly in a neural network or more explicitly in simulator. Beyond reducing the amount of training data and potentially improving the level of performance reached by the DQN, it could eliminate the need to retrain a Frostbite network if the objects (e.g., birds, ice-floes and fish) are slightly altered in their behavior, reward-structure, or appearance. When a new object type such as a bear is introduced, as in the later levels of Frostbite (Figure \ref{frostbite}D), a network endowed with intuitive physics would also have an easier time adding this object type to its knowledge \citep[the challenge of adding new objects was also discussed in][]{Marcus1998,marcus01}. In this way, the integration of intuitive physics and deep learning could be an important step towards more human-like learning algorithms.

\subsubsection{Intuitive psychology} \label{sec:psychology}

Intuitive psychology is another early-emerging ability with an important influence on human learning and thought. Pre-verbal infants distinguish animate agents from inanimate objects. This distinction is partially based on innate or early-present detectors for low-level cues, such as the presence of eyes, motion initiated from rest, and biological motion \citep{Johnson1998,Schlottmann2006,Premack1997,Tremoulet2000}. Such cues are often sufficient but not necessary for the detection of agency.

Beyond these low-level cues, infants also expect agents to act contingently and reciprocally, to have goals, and to take efficient actions towards those goals subject to constraints \citep{Spelke2007,Csibra2003,Csibra2008}. These goals can be socially directed; at around three months of age, infants begin to discriminate anti-social agents that hurt or hinder others from neutral agents \citep{KileyHamlin2010,Hamlin2013}, and they later distinguish between anti-social, neutral, and pro-social agents \citep{hamlin2007,KileyHamlin2013a}.

It is generally agreed that infants expect agents to act in a goal-directed, efficient, and socially sensitive fashion \citep{Spelke2007}. What is less agreed on is the computational architecture that supports this reasoning and whether it includes any reference to mental states and explicit goals.

One possibility is that intuitive psychology is simply cues ``all the way down'' \citep{Schlottmann2013,Scholl2013}, though this would require more and more cues as the scenarios become more complex. Consider for example a scenario in which an agent A is moving towards a box, and an agent B moves in a way that blocks A from reaching the box. Infants and adults are likely to interpret B's behavior as `hindering' \citep{Hamlin2013}. This inference could be captured by a cue that states `if an agent's expected trajectory is prevented from completion, the blocking agent is given some negative association.'

While the cue is easily calculated, the scenario is also easily changed to necessitate a different type of cue. Suppose A was already negatively associated (a `bad guy'); acting negatively towards A could then be seen as good \citep{Hamlin2013}. Or suppose something harmful was in the box which A didn't know about. Now B would be seen as helping, protecting, or defending A. Suppose A knew there was something bad in the box and wanted it anyway. B could be seen as acting paternalistically. A cue-based account would be twisted into gnarled combinations such as `If an expected trajectory is prevented from completion, the blocking agent is given some negative association, unless that trajectory leads to a negative outcome or the blocking agent is previously associated as positive, or the blocked agent is previously associated as negative, or...' 

One alternative to a cue-based account is to use generative models of action choice, as in the Bayesian inverse planning (or ``Bayesian theory-of-mind'') models of \citet{baker2009action} or the ``naive utility calculus'' models of \citet{jara2015children} (See also \citet{jern2015decision} and \citet{tauber2011using}, and a related alternative based on predictive coding from \citet{Kilner2007}). These models formalize explicitly mentalistic concepts such as `goal,' `agent,' `planning,' `cost,' `efficiency,' and `belief,' used to describe core psychological reasoning in infancy. They assume adults and children treat agents as approximately rational planners who choose the most efficient means to their goals.  Planning computations may be formalized as solutions to Markov Decision Processes (or POMDPs), taking as input utility and belief functions defined over an agent's state-space and the agent's state-action transition functions, and returning a series of actions the agent should perform to most efficiently fulfill their goals (or maximize their utility).  By simulating these planning processes, people can predict what agents might do next, or use inverse reasoning from observing a series of actions to infer the utilities and beliefs of agents in a scene.  This is directly analogous to how simulation engines can be used for intuitive physics, to predict what will happen next in a scene or to infer objects' dynamical properties from how they move.  It yields similarly flexible reasoning abilities: Utilities and beliefs can be adjusted to take into account how agents might act for a wide range of novel goals and situations.   Importantly, unlike in intuitive physics, simulation-based reasoning in intuitive psychology can be nested recursively to understand social interactions -- we can think about agents thinking about other agents. 

As in the case of intuitive physics, the success that generic deep networks will have in capturing intuitive psychological reasoning will depend in part on the representations humans use. Although deep networks have not yet been applied to scenarios involving theory-of-mind and intuitive psychology, they could probably learn visual cues, heuristics and summary statistics of a scene that happens to involve agents.\footnote{While connectionist networks have been used to model the general transition that children undergo between the ages of 3 and 4 regarding false belief \citep[e.g.,][]{berthiaume2013constructivist}, we are referring here to scenarios which require inferring goals, utilities, and relations.} If that is all that underlies human psychological reasoning, a data-driven deep learning approach can likely find success in this domain.

However, it seems to us that any full formal account of intuitive psychological reasoning needs to include representations of agency, goals, efficiency, and reciprocal relations. As with objects and forces, it is unclear whether a complete representation of these concepts (agents, goals, etc.) could emerge from deep neural networks trained in a purely predictive capacity. Similar to the intuitive physics domain, it is possible that with a tremendous number of training trajectories in a variety of scenarios, deep learning techniques could approximate the reasoning found in infancy even without learning anything about goal-directed or social-directed behavior more generally. But this is also unlikely to resemble how humans learn, understand, and apply intuitive psychology unless the concepts are genuine. In the same way that altering the setting of a scene or the target of inference in a physics-related task may be difficult to generalize without an understanding of objects, altering the setting of an agent or their goals and beliefs is difficult to reason about without understanding intuitive psychology.

In introducing the Frostbite challenge, we discussed how people can learn to play the game extremely quickly by watching an experienced player for just a few minutes and then playing a few rounds themselves. Intuitive psychology provides a basis for efficient learning from others, especially in teaching settings with the goal of communicating knowledge efficiently \citep{shafto14}. In the case of watching an expert play Frostbite, whether or not there is an explicit goal to teach, intuitive psychology lets us infer the beliefs, desires, and intentions of the experienced player. For instance, we can learn that the birds are to be avoided from seeing how the experienced player appears to avoid them. We do not need to experience a single example of encountering a bird -- and watching the Frostbite Bailey die because of the bird -- in order to infer that birds are probably dangerous. It is enough to see that the experienced player's avoidance behavior is best explained as acting under that belief.

Similarly, consider how a sidekick agent (increasingly popular in video-games) is expected to help a player achieve their goals. This agent can be useful in different ways under different circumstances, such as getting items, clearing paths, fighting, defending, healing, and providing information -- all under the general notion of being helpful \citep{macindoe2013sidekick}. An explicit agent representation can predict how such an agent will be helpful in new circumstances, while a bottom-up pixel-based representation is likely to struggle.

There are several ways that intuitive psychology could be incorporated into contemporary deep learning systems. While it could be built in, intuitive psychology may arise in other ways. Connectionists have argued that innate constraints in the form of hard-wired cortical circuits are unlikely \citep{elman2005connectionist,elman1996rethinking}, but a simple inductive bias, for example the tendency to notice things that move other things, can bootstrap reasoning about more abstract concepts of agency \citep{ullman2012simple}.\footnote{We must be careful here about what ``simple'' means. An inductive bias may appear simple in the sense that we can compactly describe it, but it may require complex computation (e.g., motion analysis, parsing images into objects, etc.) just to produce its inputs in a suitable form.} Similarly, a great deal of goal-directed and socially-directed actions can also be boiled down to a simple utility-calculus \citep[e.g.,][]{jara2015children}, in a way that could be shared with other cognitive abilities. While the origins of intuitive psychology is still a matter of debate, it is clear that these abilities are early-emerging and play an important role in human learning and thought, as exemplified in the Frostbite challenge and when learning to play novel video games more broadly.

\subsection{Learning as rapid model building}
\label{sec:learning}

Since their inception, neural networks models have stressed the importance of learning. There are many learning algorithms for neural networks, including the perceptron algorithm \citep{rosenblatt58}, Hebbian learning \citep{hebb49}, the BCM rule \citep{Bienenstock1982}, backpropagation \citep{Rumelhart1986a}, the wake-sleep algorithm \citep{Hinton1995}, and contrastive divergence \citep{Hinton2002}. Whether the goal is supervised or unsupervised learning, these algorithms implement learning as a process of gradual adjustment of connection strengths. For supervised learning, the updates are usually aimed at improving the algorithm's pattern recognition capabilities. For unsupervised learning, the updates work towards gradually matching the statistics of the model's internal patterns with the statistics of the input data.

In recent years, machine learning has found particular success using backpropagation and large data sets to solve difficult pattern recognition problems. While these algorithms have reached human-level performance on several challenging benchmarks, they are still far from matching human-level learning in other ways. Deep neural networks often need more data than people do in order to solve the same types of problems, whether it is learning to recognize a new type of object or learning to play a new game. 
When learning the meanings of words in their native language, children make meaningful generalizations from very sparse data \citep[][although see \citealt{Horst2008} regarding memory limitations]{CareyBartlett1978,Landau1988,Markman1989,smith-etal02,Xu2007}. Children may only need to see a few examples of the concepts \emph{hairbrush}, \emph{pineapple} or \emph{lightsaber} before they largely `get it,' grasping the boundary of the infinite set that defines each concept from the infinite set of all possible objects. Children are far more practiced than adults at learning new concepts -- learning roughly nine or ten new words each day after beginning to speak through the end of high school \citep{Carey1978,Bloom2000} -- yet the ability for rapid ``one-shot'' learning does not disappear in adulthood. An adult may need to see a single image or movie of a novel two-wheeled vehicle to infer the boundary between this concept and others, allowing him or her to discriminate new examples of that concept from similar looking objects of a different type (Fig. \ref{fig_characters_challenge}B-i).

Contrasting with the efficiency of human learning, neural networks -- by virtue of their generality as highly flexible function approximators -- are notoriously data hungry \citep[the bias/variance dilemma; ][]{Geman1992}. Benchmark tasks such as the ImageNet data set for object recognition provides hundreds or thousands of examples per class \citep{Krizhevsky2012,Russakovsky2014} -- 1000 hairbrushes, 1000 pineapples, etc. In the context of learning new handwritten characters or learning to play Frostbite, the MNIST benchmark includes 6000 examples of each handwritten digit \citep{LeCunBottou1998}, and the DQN of \citet{Mnih2015} played each Atari video game for approximately 924 hours of unique training experience (Figure \ref{frostbite_learning}). In both cases, the algorithms are clearly using information less efficiently than a person learning to perform the same tasks.

It is also important to mention that there are many classes of concepts that people learn more slowly. Concepts that are learned in school are usually far more challenging and more difficult to acquire, including mathematical functions, logarithms, derivatives, integrals, atoms, electrons, gravity, DNA, evolution, etc. There are also domains for which machine learners outperform human learners, such as combing through financial or weather data. But for the vast majority of cognitively natural concepts -- the types of things that children learn as the meanings of words -- people are still far better learners than machines. This is the type of learning we focus on in this section, which is more suitable for the enterprise of reverse engineering and articulating additional principles that make human learning successful. It also opens the possibility of building these ingredients into the next generation of machine learning and AI algorithms, with potential for making progress on learning concepts that are both easy and difficult for humans to acquire.

Even with just a few examples, people can learn remarkably rich conceptual models. One indicator of richness is the variety of functions that these models support \citep{Solomon1999,Markman2003}. Beyond classification, concepts support prediction \citep{Rips1975,Murphy1994}, action \citep{Barsalou1983}, communication \citep{Markman1998}, imagination \citep{Ward1994,Jern2013}, explanation \citep{Lombrozo2009,Williams2010}, and composition \citep{Osherson1981,Murphy1988}. These abilities are not independent; rather they hang together and interact \citep{Solomon1999}, coming for free with the acquisition of the underlying concept. Returning to the previous example of a novel two wheeled vehicle, a person can sketch a range of new instances (Figure \ref{fig_characters_challenge}B-ii), parse the concept into its most important components (Figure \ref{fig_characters_challenge}B-iii), or even create a new complex concept through the combination of familiar concepts (Figure \ref{fig_characters_challenge}B-iv). Likewise, as discussed in the context of Frostbite, a learner who has acquired the basics of the game could flexibly apply their knowledge to an infinite set of Frostbite variants (Section \ref{frostbite_section}). The acquired knowledge supports reconfiguration to new tasks and new demands, such as modifying the goals of the game to survive while acquiring as few points as possible, or to efficiently teach the rules to a friend.

This richness and flexibility suggests that learning as model building is a better metaphor than learning as pattern recognition. Furthermore, the human capacity for one-shot learning suggests that these models are built upon rich domain knowledge rather than starting from a blank slate \citep{Mitchell1986,Mikolov2015}. In contrast, much of the recent progress in deep learning has been on pattern recognition problems, including object recognition, speech recognition, and (model-free) video game learning, that utilize large data sets and little domain knowledge.

There has been recent work on other types of tasks including learning generative models of images \citep{Gregor2014,Denton2015a}, caption generation \citep{Vinyals2014,Karpathy2015,Xu2015}, question answering \citep{Weston2015,Sukhbaatar2015}, and learning simple algorithms \citep{Graves2014,Grefenstette2015}; we discuss question answering and learning simple algorithms in Section \ref{sec:directions}. Yet, at least for image and caption generation, these tasks have been mostly studied in the big data setting that is at odds with the impressive human ability for generalizing from small data sets \citep[although see][for a deep learning approach to the Character Challenge]{Rezende2016}. And it has been difficult to learn neural-network-style representations that effortlessly generalize to new tasks that they were not trained on \citep[see][]{Marcus1998,marcus01,Davis2015}. What additional ingredients may be needed in order to rapidly learn more powerful and more general-purpose representations?

\begin{figure}
\centering\includegraphics[width=6.5in]{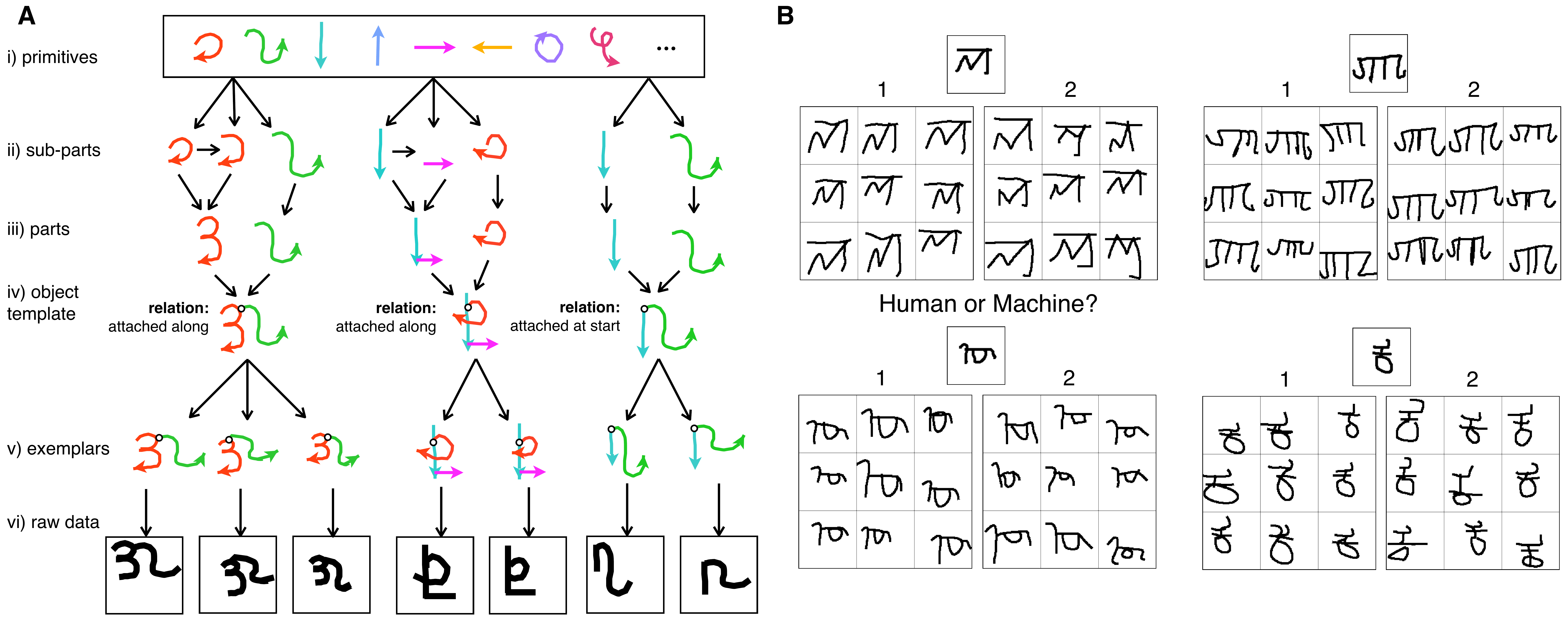}
\caption{A causal, compositional model of handwritten characters. A) New types are generated compositionally by choosing primitive actions (color coded) from a library (i), combining these sub-parts (ii) to make parts (iii), and combining parts with relations to define simple programs (iv). These programs can create different tokens of a concept (v) that are rendered as binary images (vi). B) Probabilistic inference allows the model to generate new examples from just one example of a new concept, shown here in a visual Turing Test. An example image of a new concept is shown above each pair of grids. One grid was generated by 9 people and the other is 9 samples from the BPL model. Which grid in each pair (A or B) was generated by the machine? Answers by row: 1,2;1,1. Adapted from \citet{LakeScience2015}.}
\label{BPL}
\end{figure}

A relevant case study is from our own work on the Characters Challenge \citep[Section \ref{characters_section}; ][]{LakeScience2015,Lake2014a}. People and various machine learning approaches were compared on their ability to learn new handwritten characters from the world's alphabets. In addition to evaluating several types of deep learning models, we developed an algorithm using Bayesian Program Learning (BPL) that represents concepts as simple stochastic programs -- that is, structured procedures that generate new examples of a concept when executed (Figure \ref{BPL}A). These programs allow the model to express causal knowledge about how the raw data are formed, and the probabilistic semantics allow the model to handle noise and perform creative tasks. Structure sharing across concepts is accomplished by the compositional reuse of stochastic primitives that can combine in new ways to create new concepts.

Note that we are overloading the word ``model'' to refer to both the BPL framework as a whole (which is a generative model), as well as the individual probabilistic models (or concepts) that it infers from images to represent novel handwritten characters. There is a hierarchy of models: a higher-level program that generates different types of concepts, which are themselves programs that can be run to generate tokens of a concept. Here, describing learning as ``rapid model building'' refers to the fact that BPL constructs generative models (lower-level programs) that produce tokens of a concept (Figure \ref{BPL}B). 

Learning models of this form allows BPL to perform a challenging one-shot classification task at human level performance (Figure \ref{fig_characters_challenge}A-i) and to outperform current deep learning models such as convolutional networks \citep{Koch2015}.\footnote{A new approach using convolutional ``matching networks'' achieves good one-shot classification performance when discriminating between characters from different alphabets \citep{Vinyals2016}. It has not yet been directly compared with BPL, which was evaluated on one-shot classification with characters from the same alphabet.} The representations that BPL learns also enable it to generalize in other, more creative human-like ways, as evaluated using ``visual Turing tests" (e.g., Figure \ref{BPL}B). These tasks include generating new examples (Figure \ref{fig_characters_challenge}A-ii and Figure \ref{BPL}B), parsing objects into their essential components (Figure \ref{fig_characters_challenge}A-iii), and generating new concepts in the style of a particular alphabet (Figure \ref{fig_characters_challenge}A-iv). The following sections discuss the three main ingredients -- compositionality, causality, and learning-to-learn -- that were important to the success of this framework and we believe are important to understanding human learning as rapid model building more broadly. While these ingredients fit naturally within a BPL or a probabilistic program induction framework, they could also be integrated into deep learning models and other types of machine learning algorithms, prospects we discuss in more detail below.

\subsubsection{Compositionality} \label{sec:compositionality}

Compositionality is the classic idea that new representations can be constructed through the combination of primitive elements. In computer programming, primitive functions can be combined together to create new functions, and these new functions can be further combined to create even more complex functions. This function hierarchy provides an efficient description of higher-level functions, like a part hierarchy for describing complex objects or scenes \citep{Bienenstock1997}. Compositionality is also at the core of productivity: an infinite number of representations can be constructed from a finite set of primitives, just as the mind can think an infinite number of thoughts, utter or understand an infinite number of sentences, or learn new concepts from a seemingly infinite space of possibilities \citep{Fodor1975,fodor88,marcus01,Piantadosi2011}.

Compositionality has been broadly influential in both AI and cognitive science, especially as it pertains to theories of object recognition, conceptual representation, and language. Here we focus on compositional representations of object concepts for illustration. Structural description models represent visual concepts as compositions of parts and relations, which provides a strong inductive bias for constructing models of new concepts \citep{Marr1978,Winston1975,Biederman1987,Hummel1992,VandenHengel2015}. For instance, the novel two-wheeled vehicle in Figure \ref{fig_characters_challenge}B might be represented as two wheels connected by a platform, which provides the base for a post, which holds the handlebars, etc. Parts can themselves be composed of sub-parts, forming a ``partonomy'' of part-whole relationships \citep{Miller1976,Tversky1984}. In the novel vehicle example, the parts and relations can be shared and reused from existing related concepts, such as cars, scooters, motorcycles, and unicycles. Since the parts and relations are themselves a product of previous learning, their facilitation of the construction of new models is also an example of learning-to-learn -- another ingredient that is covered below. While compositionality and learning-to-learn fit naturally together, there are also forms of compositionality that rely less on previous learning, such as the bottom-up parts-based representation of \citet{Hoffman1984}.

Learning models of novel handwritten characters can be operationalized in a similar way. Handwritten characters are inherently compositional, where the parts are pen strokes and relations describe how these strokes connect to each other. \citet{LakeScience2015} modeled these parts using an additional layer of compositionality, where parts are complex movements created from simpler sub-part movements. New characters can be constructed by combining parts, sub-parts, and relations in novel ways (Figure \ref{BPL}). Compositionality is also central to the construction of other types of symbolic concepts beyond characters, where new spoken words can be created through a novel combination of phonemes \citep{Lake2014} or a new gesture or dance move can be created through a combination of more primitive body movements.

An efficient representation for Frostbite should be similarly compositional and productive. A scene from the game is a composition of various object types, including birds, fish, ice floes, igloos, etc. (Figure \ref{frostbite}). Representing this compositional structure explicitly is both more economical and better for generalization, as noted in previous work on object-oriented reinforcement learning \citep{Diuk2008}. Many repetitions of the same objects are present at different locations in the scene, and thus representing each as an identical instance of the same object with the same properties is important for efficient representation and quick learning of the game. Further, new levels may contain different numbers and combinations of objects, where a compositional representation of objects -- using intuitive physics and intuitive psychology as glue -- would aid in making these crucial generalizations (Figure \ref{frostbite}D).

Deep neural networks have at least a limited notion of compositionality. Networks trained for object recognition encode part-like features in their deeper layers \citep{Zeiler2012}, whereby the presentation of new types of objects can activate novel combinations of feature detectors. Similarly, a DQN trained to play Frostbite may learn to represent multiple replications of the same object with the same features, facilitated by the invariance properties of a convolutional neural network architecture. Recent work has shown how this type of compositionality can be made more explicit, where neural networks can be used for efficient inference in more structured generative models (both neural networks and 3D scene models) that explicitly represent the number of objects in a scene \citep{eslami16}. Beyond the compositionality inherent in parts, objects, and scenes, compositionality can also be important at the level of goals and sub-goals. Recent work on hierarchical-DQNs shows that by providing explicit object representations to a DQN, and then defining sub-goals based on reaching those objects, DQNs can learn to play games with sparse rewards (such as Montezuma's Revenge) by combining these sub-goals together to achieve larger goals \citep{Kulkarni2016}.

We look forward to seeing these new ideas continue to develop, potentially providing even richer notions of compositionality in deep neural networks that lead to faster and more flexible learning. To capture the full extent of the mind's compositionality, a model must include explicit representations of objects, identity, and relations -- all while maintaining a notion of ``coherence'' when understanding novel configurations. Coherence is related to our next principle, causality, which is discussed in the section that follows.

\begin{figure}
\centering\includegraphics[width=6.75in]{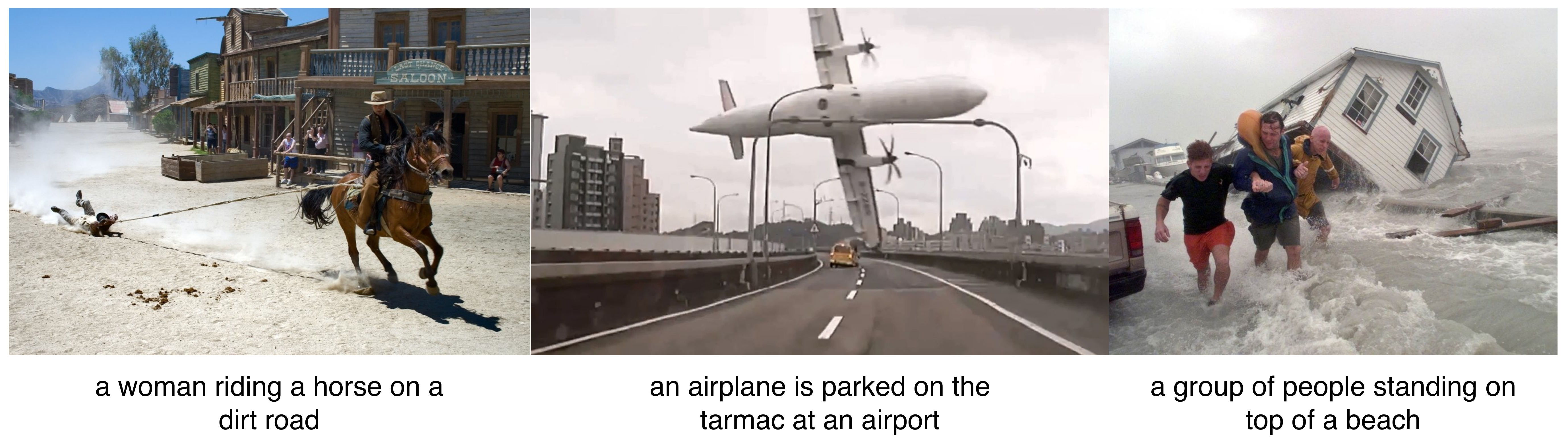}
\caption{Perceiving scenes without intuitive physics, intuitive psychology, compositionality, and causality. Image captions are generated by a deep neural network \citep{Karpathy2015} using code from \url{github.com/karpathy/neuraltalk2}. Image credits: Gabriel Villena Fern\'{a}ndez (left), TVBS Taiwan / Agence France-Presse (middle) and AP Photo / Dave Martin (right). Similar examples using images from Reuters news can be found at \url{twitter.com/interesting\_jpg}. }
\label{caption_generation}
\end{figure}

\subsubsection{Causality} \label{sec:causality}
In concept learning and scene understanding, causal models represent hypothetical real world processes that produce the perceptual observations. In control and reinforcement learning, causal models represent the structure of the environment, such as modeling state-to-state transitions or action/state-to-state transitions.

Concept learning and vision models that utilize causality are usually generative (as opposed to discriminative; see Glossary in Table \ref{glossary}), but not every generative model is also causal. While a generative model describes a process for generating data, or at least assigns a probability distribution over possible data points, this generative process may not resemble how the data are produced in the real world. Causality refers to the subclass of generative models that resemble, at an abstract level, how the data are actually generated. While generative neural networks such as Deep Belief Networks \citep{Hinton2006a} or variational auto-encoders \citep{Kingma2014b,Gregor2016} may generate compelling handwritten digits, they mark one end of the ``causality spectrum,'' since the steps of the generative process bear little resemblance to steps in the actual process of writing. In contrast, the generative model for characters using Bayesian Program Learning (BPL) does resemble the steps of writing, although even more causally faithful models are possible.

Causality has been influential in theories of perception. ``Analysis-by-synthesis'' theories of perception maintain that sensory data can be more richly represented by modeling the process that generated it \citep{Halle1962,Eden1962,Neisser,Bever2010}. Relating data to its causal source provides strong priors for perception and learning, as well as a richer basis for generalizing in new ways and to new tasks. The canonical examples of this approach are speech and visual perception. For instance, \citet{Liberman1967} argued that the richness of speech perception is best explained by inverting the production plan, at the level of vocal tract movements, in order to explain the large amounts of acoustic variability and the blending of cues across adjacent phonemes. As discussed, causality does not have to be a literal inversion of the actual generative mechanisms, as proposed in the motor theory of speech. For the BPL of learning handwritten characters, causality is operationalized by treating concepts as motor programs, or abstract causal descriptions of how to produce examples of the concept, rather than concrete configurations of specific muscles (Figure \ref{BPL}A). Causality is an important factor in the model's success in classifying and generating new examples after seeing just a single example of a new concept \citep{LakeScience2015} (Figure \ref{BPL}B).

Causal knowledge has also been shown to influence how people learn new concepts; providing a learner with different types of causal knowledge changes how they learn and generalize. For example, the structure of the causal network underlying the features of a category influences how people categorize new examples \citep{Rehder2001,Rehder2003}. Similarly, as related to the Characters Challenge, the way people learn to write a novel handwritten character influences later perception and categorization \citep{Freyd1983,Freyd1987}.

To explain the role of causality in learning, conceptual representations have been likened to intuitive theories or explanations, providing the glue that lets core features stick while other equally applicable features wash away \citep{Murphy1985}. Borrowing examples from \citet{Murphy1985}, the feature ``flammable'' is more closely attached to wood than money due to the underlying causal roles of the concepts, even though the feature is equally applicable to both; these causal roles derive from the \emph{functions} of objects. Causality can also glue some features together by relating them to a deeper underlying cause, explaining why some features such as ``can fly,'' ``has wings,'' and ``has feathers'' co-occur across objects while others do not.

Beyond concept learning, people also understand scenes by building causal models. Human-level scene understanding involves composing a story that explains the perceptual observations, drawing upon and integrating the ingredients of intuitive physics, intuitive psychology, and compositionality. Perception without these ingredients, and absent the causal glue that binds them together, can lead to revealing errors. Consider image captions generated by a deep neural network \citep[Figure \ref{caption_generation}; ][]{Karpathy2015}. In many cases, the network gets the key objects in a scene correct but fails to understand the physical forces at work, the mental states of the people, or the causal relationships between the objects -- in other words, it does not build the right causal model of the data.

There have been steps towards deep neural networks and related approaches that learn causal models. \citet{Lopez-paz2015} introduced a discriminative, data-driven framework for distinguishing the direction of causality from examples. While it outperforms existing methods on various causal prediction tasks, it is unclear how to apply the approach to inferring rich hierarchies of latent causal variables, as needed for the Frostbite Challenge and (especially) the Characters Challenge. \citet{Graves2013} learned a generative model of cursive handwriting using a recurrent neural network trained on handwriting data. While it synthesizes impressive examples of handwriting in various styles, it requires a large training corpus and has not been applied to other tasks. The DRAW network performs both recognition and generation of handwritten digits using recurrent neural networks with a window of attention, producing a limited circular area of the image at each time step \citep{Gregor2014}. A more recent variant of DRAW was applied to generating examples of a novel character from just a single training example \citep{Rezende2016}. While the model demonstrates an impressive ability to make plausible generalizations that go beyond the training examples, it generalizes too broadly in other cases, in ways that are not especially human-like.  It is not clear that it could yet pass any of the ``visual Turing tests'' in \citet{LakeScience2015} (Figure \ref{BPL}B), although we hope DRAW-style networks will continue to be extended and enriched, and could be made to pass these tests.

Incorporating causality may greatly improve these deep learning models; they were trained without access to causal data about how characters are actually produced, and without any incentive to learn the true causal process.  An attentional window is only a crude approximation to the true causal process of drawing with a pen, and in \citet{Rezende2016} the attentional window is not pen-like at all, although a more accurate pen model could be incorporated. We anticipate that these sequential generative neural networks could make sharper one-shot inferences -- with the goal of tackling the full Characters Challenge -- by incorporating additional causal, compositional, and hierarchical structure (and by continuing to utilize learning-to-learn, described next), potentially leading to a more computationally efficient and neurally grounded variant of the BPL model of handwritten characters (Figure \ref{BPL}).

A causal model of Frostbite would have to be more complex, gluing together object representations and explaining their interactions with intuitive physics and intuitive psychology, much like the game engine that generates the game dynamics and ultimately the frames of pixel images. Inference is the process of inverting this causal generative model, explaining the raw pixels as objects and their interactions, such as the agent stepping on an ice floe to deactivate it or a crab pushing the agent into the water (Figure \ref{frostbite}). Deep neural networks could play a role in two ways: serving as a bottom-up proposer to make probabilistic inference more tractable in a structured generative model (Section \ref{sec:inference}) or by serving as the causal generative model if imbued with the right set of ingredients.

\subsubsection{Learning-to-learn} \label{sec:L2L}
When humans or machines make inferences that go far beyond the data, strong prior knowledge (or inductive biases or constraints) must be making up the difference \citep{Geman1992,Griffiths2010a,Tenenbaum2011}. One way people acquire this prior knowledge is through ``learning-to-learn,'' a term introduced by \citet{Harlow1949} and closely related to the machine learning notions of ``transfer learning'', ``multi-task learning'' or ``representation learning.'' These terms refer to ways that learning a new task (or a new concept) can be accelerated through previous or parallel learning of other related tasks (or other related concepts). The strong priors, constraints, or inductive bias needed to learn a particular task quickly are often shared to some extent with other related tasks. A range of mechanisms have been developed to adapt the learner's inductive bias as they learn specific tasks, and then apply these inductive biases to new tasks.

In hierarchical Bayesian modeling \citep{Gelman2004}, a general prior on concepts is shared by multiple specific concepts, and the prior itself is learned over the course of learning the specific concepts \citep{Salakhutdinov2012,Salakhutdinov2013}. These models have been used to explain the dynamics of human learning-to-learn in many areas of cognition, including word learning, causal learning, and learning intuitive theories of physical and social domains \citep{Tenenbaum2011}.  In machine vision, for deep convolutional networks or other discriminative methods that form the core of recent recognition systems, learning-to-learn can occur through the sharing of features between the models learned for old objects (or old tasks) and the models learned for new objects (or new tasks) \citep{Baxter2000,TorralbaMurphy2007,bottou2014,Salakhutdinov2011a,Srivastava2013,Zeiler2012,Anselmi2015,Lopez-paz2016,Rusu2016}. Neural networks can also learn-to-learn by optimizing hyperparameters, including the form of their weight update rule \citep{Andrychowicz2016}, over a set of related tasks.

While transfer learning and multi-task learning are already important themes across AI, and in deep learning in particular, they have not yet led to systems that learn new tasks as rapidly and flexibly as humans do. Capturing more human-like learning-to-learn dynamics in deep networks and other machine learning approaches could facilitate much stronger transfer to new tasks and new problems. To gain the full benefit that humans get from learning-to-learn, however, AI systems might first need to adopt the more compositional (or more language-like, see Section \ref{sec:objections}) and causal forms of representations that we have argued for above.

We can see this potential in both of our Challenge problems. In the Characters Challenge as presented in \citet{LakeScience2015}, all viable models use ``pre-training'' on many character concepts in a background set of alphabets to tune the representations they use to learn new character concepts in a test set of alphabets. But to perform well, current neural network approaches require much more pre-training than do people or our Bayesian program learning approach, and they are still far from solving the Characters Challenge.\footnote{Humans typically have direct experience with only one or a few alphabets, and even with related drawing experience, this likely amounts to the equivalent of a few hundred character-like visual concepts at most. For BPL, pre-training with characters in only five alphabets (for around 150 character types in total) is sufficient to perform human-level one-shot classification and generation of new examples. The best neural network classifiers (deep convolutional networks) have error rates approximately five times higher than humans when pre-trained with five alphabets ($23\%$ versus $4\%$ error), and two to three times higher when pre-training on six times as much data (30 alphabets) \citep{LakeScience2015}. The current need for extensive pre-training is illustrated for deep generative models by \citet{Rezende2016}, who present extensions of the DRAW architecture capable of one-shot learning.}

We cannot be sure how people get to the knowledge they have in this domain, but we do understand how this works in BPL, and we think people might be similar. BPL transfers readily to new concepts because it learns about object parts, sub-parts, and relations, capturing learning about what each concept is like and what concepts are like in general. It is crucial that learning-to-learn occurs at multiple levels of the hierarchical generative process. Previously learned primitive actions and larger generative pieces can be re-used and re-combined to define new generative models for new characters (Figure \ref{BPL}A). Further transfer occurs by learning about the typical levels of variability within a typical generative model; this provides knowledge about how far and in what ways to generalize when we have seen only one example of a new character, which on its own could not possibly carry any information about variance. BPL could also benefit from deeper forms of learning-to-learn than it currently does: Some of the important structure it exploits to generalize well is built in to the prior and not learned from the background pre-training, whereas people might learn this knowledge, and ultimately a human-like machine learning system should as well.

Analogous learning-to-learn occurs for humans in learning many new object models, in vision and cognition: Consider the novel two-wheeled vehicle in Figure \ref{fig_characters_challenge}B, where learning-to-learn can operate through the transfer of previously learned parts and relations (sub-concepts such as wheels, motors, handle bars, attached, powered by, etc.) that reconfigure compositionally to create a model of the new concept.  If deep neural networks could adopt similarly compositional, hierarchical, and causal representations, we expect they might benefit more from learning-to-learn.

In the Frostbite Challenge, and in video games more generally, there is a similar interdependence between the form of the representation and the effectiveness of learning-to-learn. People seem to transfer knowledge at multiple levels, from low-level perception to high-level strategy, exploiting compositionality at all levels.  Most basically, they immediately parse the game environment into objects, types of objects, and causal relations between them. People also understand that video games like this have goals, which often involve approaching or avoiding objects based on their type. Whether the person is a child or a seasoned gamer, it seems obvious that interacting with the birds and fish will change the game state in some way, either good or bad, because video games typically yield costs or rewards for these types of interactions (e.g., dying or points). These types of hypotheses can be quite specific and rely on prior knowledge: When the polar bear first appears and tracks the agent's location during advanced levels (Figure \ref{frostbite}D), an attentive learner is sure to avoid it. Depending on the level, ice floes can be spaced far apart (Figure \ref{frostbite}A-C) or close together (Figure \ref{frostbite}D), suggesting the agent may be able to cross some gaps but not others. In this way, general world knowledge and previous video games may help inform exploration and generalization in new scenarios, helping people learn maximally from a single mistake or avoid mistakes altogether.

Deep reinforcement learning systems for playing Atari games have had some impressive successes in transfer learning, but they still have not come close to learning to play new games as quickly as humans can. For example, \citet{Parisotto2015} presents the ``Actor-mimic'' algorithm that first learns 13 Atari games by watching an expert network play and trying to mimic the expert network action selection and/or internal states (for about four million frames of experience each, or 18.5 hours per game). This algorithm can then learn new games faster than a randomly initialized DQN: Scores that might have taken four or five million frames of learning to reach might now be reached after one or two million frames of practice. But anecdotally we find that humans can still reach these scores with a few minutes of practice, requiring far less experience than the DQNs.

In sum, the interaction between representation and previous experience may be key to building machines that learn as fast as people do. A deep learning system trained on many video games may not, by itself, be enough to learn new games as quickly as people do. Yet if such a system aims to learn compositionally structured causal models of a each game -- built on a foundation of intuitive physics and psychology -- it could transfer knowledge more efficiently and thereby learn new games much more quickly.

\subsection{Thinking Fast} \label{sec:thinkingfast}
The previous section focused on learning rich models from sparse data and proposed ingredients for achieving these human-like learning abilities. These cognitive abilities are even more striking when considering the speed of perception and thought -- the amount of time required to understand a scene, think a thought, or choose an action. In general, richer and more structured models require more complex (and slower) inference algorithms -- similar to how complex models require more data -- making the speed of perception and thought all the more remarkable.

The combination of rich models with efficient inference suggests another way psychology and neuroscience may usefully inform AI. It also suggests an additional way to build on the successes of deep learning, where efficient inference and scalable learning are important strengths of the approach. This section discusses possible paths towards resolving the conflict between fast inference and structured representations, including Helmholtz-machine-style approximate inference in generative models \citep{Hinton1995,Dayan1995} and cooperation between model-free and model-based reinforcement learning systems.

\subsubsection{Approximate inference in structured models} \label{sec:inference}

Hierarchical Bayesian models operating over probabilistic programs \citep{Goodman2008,Tenenbaum2011,LakeScience2015} are equipped to deal with theory-like structures and rich causal representations of the world, yet there are formidable algorithmic challenges for efficient inference. Computing a probability distribution over an entire space of programs is usually intractable, and often even finding a single high-probability program poses an intractable search problem. In contrast, while representing intuitive theories and structured causal models is less natural in deep neural networks, recent progress has demonstrated the remarkable effectiveness of gradient-based learning in high-dimensional parameter spaces. A complete account of learning and inference must explain how the brain does so much with limited computational resources \citep{gershman15,Vul2014}.

Popular algorithms for approximate inference in probabilistic machine learning have been proposed as psychological models \citep[see][for a review]{griffiths12}. Most prominently, it has been proposed that humans can approximate Bayesian inference using Monte Carlo methods, which stochastically sample the space of possible hypotheses and evaluate these samples according to their consistency with the data and prior knowledge \citep{bonawitz14,gershman12,Ullman2012,Vul2014}. Monte Carlo sampling has been invoked to explain behavioral phenomena ranging from children's response variability \citep{bonawitz14} to garden-path effects in sentence processing \citep{levy09} and perceptual multistability \citep{gershman12,moreno11}. Moreover, we are beginning to understand how such methods could be implemented in neural circuits \citep{buesing11,huang14,pecevski11}.\footnote{In the interest of brevity, we do not discuss here another important vein of work linking neural circuits to variational approximations \citep{bastos12}, which have received less attention in the psychological literature.}

While Monte Carlo methods are powerful and come with asymptotic guarantees, it is challenging to make them work on complex problems like program induction and theory learning. When the hypothesis space is vast and only a few hypotheses are consistent with the data, how can good models be discovered without exhaustive search? In at least some domains, people may not have an especially clever solution to this problem, instead grappling with the full combinatorial complexity of theory learning \citep{Ullman2012}. Discovering new theories can be slow and arduous, as testified by the long timescale of cognitive development, and learning in a saltatory fashion (rather than through gradual adaptation) is characteristic of aspects of human intelligence, including discovery and insight during development \citep{Schulz2012}, problem-solving \citep{sternberg95}, and epoch-making discoveries in scientific research \citep{langley87}. Discovering new theories can also happen much more quickly -- A person learning the rules of Frostbite will probably undergo a loosely ordered sequence of ``Aha!'' moments: they will learn that jumping on ice floes causes them to change color, changing the color of ice floes causes an igloo to be constructed piece-by-piece, that birds make you lose points, that fish make you gain points, that you can change the direction of ice floe at the cost of one igloo piece, and so on. These little fragments of a ``Frostbite theory'' are assembled to form a causal understanding of the game relatively quickly, in what seems more like a guided process than arbitrary proposals in a Monte Carlo inference scheme. Similarly, as described in the Characters Challenge, people can quickly infer motor programs to draw a new character in a similarly guided processes.

For domains where program or theory learning happens quickly, it is possible that people employ inductive biases not only to evaluate hypotheses, but also to guide hypothesis selection. \citet{Schulz2012} has suggested that abstract structural properties of problems contain information about the abstract forms of their solutions. Even without knowing the answer to the question ``Where is the deepest point in the Pacific Ocean?'' one still knows that the answer must be a location on a map. The answer ``20 inches'' to the question ``What year was Lincoln born?'' can be invalidated \emph{a priori}, even without knowing the correct answer. In recent experiments, \citet{tsividis15} found that children can use high-level abstract features of a domain to guide hypothesis selection, by reasoning about distributional properties like the ratio of seeds to flowers, and dynamical properties like periodic or monotonic relationships between causes and effects \citep[see also][]{magid15}.

How might efficient mappings from questions to a plausible subset of answers be learned? Recent work in AI spanning both deep learning and graphical models has attempted to tackle this challenge by ``amortizing'' probabilistic inference computations into an efficient feed-forward mapping \citep{eslami14,heess13,mnih14,stuhlmuller13}.  We can also think of this as ``learning to do inference,'' which is independent from the ideas of learning as model building discussed in the previous section. These feed-forward mappings  can be learned in various ways, for example, using paired generative/recognition networks \citep{Hinton1995,Dayan1995} and variational optimization \citep{mnih14,Rezende2014,Gregor2014} or nearest-neighbor density estimation \citep{stuhlmuller13,Kulkarni2015a}. One implication of amortization is that solutions to different problems will become correlated due to the sharing of amortized computations; some evidence for inferential correlations in humans was reported by \citet{gershman14}. This trend is an avenue of potential integration of deep learning models with probabilistic models and probabilistic programming: training neural networks to help perform probabilistic inference in a generative model or a probabilistic program \citep{Yildirim,Kulkarni2015,eslami16}. Another avenue for potential integration is through differentiable programming \citep{DifferentiableProgramming} -- by ensuring that the program-like hypotheses are differentiable and thus learnable via gradient descent -- a possibility discussed in the concluding section (Section \ref{sec:directions}).

\subsubsection{Model-based and model-free reinforcement learning} \label{sec:RL-2types}
The DQN introduced by \citet{Mnih2015} used a simple form of model-free reinforcement learning in a deep neural network that allows for fast selection of actions. There is indeed substantial evidence that the brain uses similar model-free learning algorithms in simple associative learning or discrimination learning tasks \citep[see][for a review]{niv09}. In particular, the phasic firing of midbrain dopaminergic neurons is qualitatively \citep{schultz97} and quantitatively \citep{bayer05} consistent with the reward prediction error that drives updating of model-free value estimates.

Model-free learning is not, however, the whole story. Considerable evidence suggests that the brain also has a model-based learning system, responsible for building a ``cognitive map'' of the environment and using it to plan action sequences for more complex tasks \citep{daw05,dolan13}. Model-based planning is an essential ingredient of human intelligence, enabling flexible adaptation to new tasks and goals; it is where all of the rich model-building abilities discussed in the previous sections earn their value as guides to action. As we argued in our discussion of Frostbite, one can design numerous variants of this simple video game that are identical except for the reward function -- that is, governed by an identical environment model of state-action-dependent transitions.  We conjecture that a competent Frostbite player can easily shift behavior appropriately, with little or no additional learning, and it is hard to imagine a way of doing that other than having a model-based planning approach in which the environment model can be modularly combined with arbitrary new reward functions and then deployed immediately for planning. One boundary condition on this flexibility is the fact that the skills become ``habitized'' with routine application, possibly reflecting a shift from model-based to model-free control. This shift may arise from a rational arbitration between learning systems to balance the trade-off between flexibility and speed \citep{daw05,keramati11}.

Similarly to how probabilistic computations can be amortized for efficiency (see previous section), plans can be amortized into cached values by allowing the model-based system to simulate training data for the model-free system \citep{sutton90}. This process might occur offline (e.g., in dreaming or quiet wakefulness), suggesting a form of consolidation in reinforcement learning \citep{gershman14b}. Consistent with the idea of cooperation between learning systems, a recent experiment demonstrated that model-based behavior becomes automatic over the course of training \citep{economides15}. Thus, a marriage of flexibility and efficiency might be achievable if we use the human reinforcement learning systems as guidance.

Intrinsic motivation also plays an important role in human learning and behavior \citep{deci75,harlow50,berlyne66}. While much of the previous discussion assumes the standard view of behavior as seeking to maximize reward and minimize punishment, all externally provided rewards are reinterpreted according to the ``internal value'' of the agent, which may depend on the current goal and mental state. There may also be an intrinsic drive to reduce uncertainty and construct models of the environment \citep{Edelman2015a,schmidhuber15}, closely related to learning-to-learn and multi-task learning. Deep reinforcement learning is only just starting to address intrinsically motivated learning \citep{mohamed15,Kulkarni2016}.

\section{Responses to common questions} \label{sec:objections}

In discussing the arguments in this paper with colleagues, three lines of questioning or critiques have come up frequently. We think it is helpful to address these points directly, to maximize the potential for moving forward together.

\textbf{1. Comparing the learning speeds of humans and neural networks on specific tasks is not meaningful, because humans have extensive prior experience.}

It may seem unfair to compare neural networks and humans on the amount of training experience required to perform a task, such as learning to play new Atari games or learning new handwritten characters, when humans have had extensive prior experience that these networks have not benefited from. People have had many hours playing other games, and experience reading or writing many other handwritten characters, not to mention experience in a variety of more loosely related tasks.  If neural networks were ``pre-trained'' on the same experience, the argument goes, then they might generalize similarly to humans when exposed to novel tasks.

This has been the rationale behind multi-task learning or transfer learning, a strategy with a long history that has shown some promising results recently with deep networks \citep[e.g.,][]{donahue13,luong15,Parisotto2015}. Furthermore, some deep learning advocates argue, the human brain effectively benefits from even more experience through evolution. If deep learning researchers see themselves as trying to capture the equivalent of humans' collective evolutionary experience, this would be equivalent to a truly immense ``pre-training'' phase.

We agree that humans have a much richer starting point than neural networks when learning most new tasks, including learning a new concept or to play a new video game. That is the point of the ``developmental start-up software'' and other building blocks that we argued are key to creating this richer starting point. We are less committed to a particular story regarding the origins of the ingredients, including the relative roles of genetically programmed and experience-driven developmental mechanisms in building these components in early infancy.  Either way, we see them as fundamental building blocks for facilitating rapid learning from sparse data.

Learning-to-learn across multiple tasks is conceivably one route to acquiring these ingredients, but simply training conventional neural networks on many related tasks may not be sufficient to generalize in human-like ways for novel tasks. As we argued in Section \ref{sec:L2L}, successful learning-to-learn -- or at least, human-level transfer learning -- is enabled by having models with the right representational structure, including the other building blocks discussed in this paper. Learning-to-learn is a powerful ingredient, but it can be more powerful when operating over compositional representations that capture the underlying causal structure of the environment, while also building on the intuitive physics and psychology.

Finally, we recognize that some researchers still hold out hope that if only they can just get big enough training datasets, sufficiently rich tasks, and enough computing power -- far beyond what has been tried out so far -- then deep learning methods might be sufficient to learn representations equivalent to what evolution and learning provides humans with. We can sympathize with that hope and believe it deserves further exploration, although we are not sure it is a realistic one. We understand in principle how evolution could build a brain with the cognitive ingredients we discuss here. Stochastic hill-climbing is slow -- it may require massively parallel exploration, over millions of years with innumerable dead-ends -- but it can build complex structures with complex functions if we are willing to wait long enough. In contrast, trying to build these representations from scratch using backpropagation, deep Q-learning or any stochastic gradient-descent weight update rule in a fixed network architecture may be unfeasible regardless of how much training data are available. To build these representations from scratch might require exploring fundamental structural variations in the network's architecture, which gradient-based learning in weight space is not prepared to do. Although deep learning researchers do explore many such architectural variations, and have been devising increasingly clever and powerful ones recently, it is the researchers who are driving and directing this process. Exploration and creative innovation in the space of network architectures have not yet been made algorithmic. Perhaps they could, using genetic programming methods \citep{koza1992genetic} or other structure-search algorithms \citep{Yamins2014}. We think this would be a fascinating and promising direction to explore, but we may have to acquire more patience than machine learning researchers typically express with their algorithms: the dynamics of structure-search may look much more like the slow random hill-climbing of evolution than the smooth, methodical progress of stochastic gradient-descent.  An alternative strategy is to build in appropriate infant-like knowledge representations and core ingredients as the starting point for our learning-based AI systems, or to build learning systems with strong inductive biases that guide them in this direction. 

Regardless of which way an AI developer chooses to go, our main points are orthogonal to this objection. There are a set of core cognitive ingredients for human-like learning and thought.  Deep learning models could incorporate these ingredients through some combination of additional structure and perhaps additional learning mechanisms, but for the most part have yet to do so.  Any approach to human-like AI, whether based on deep learning or not, is likely to gain from incorporating these ingredients.

\textbf{2. Biological plausibility suggests theories of intelligence should start with neural networks.}

We have focused on how cognitive science can motivate and guide efforts to engineer human-like AI, in contrast to some advocates of deep neural networks who cite neuroscience for inspiration. Our approach is guided by a pragmatic view that the clearest path to a computational formalization of human intelligence comes from understanding the ``software'' before the ``hardware.'' In the case of this article, we proposed key ingredients of this software in previous sections.

Nonetheless, a cognitive approach to intelligence should not ignore what we know about the brain. Neuroscience can provide valuable inspirations for both cognitive models and AI researchers: the centrality of neural networks and model-free reinforcement learning in our proposals for ``Thinking fast'' (Section \ref{sec:thinkingfast}) are prime exemplars. Neuroscience can also in principle impose constraints on cognitive accounts, both at the cellular and systems level. If deep learning embodies brain-like computational mechanisms and those mechanisms are incompatible with some cognitive theory, then this is an argument against that cognitive theory and in favor of deep learning. Unfortunately, what we ``know'' about the brain is not all that clear-cut. Many seemingly well-accepted ideas regarding neural computation are in fact biologically dubious, or uncertain at best -- and thus should not disqualify cognitive ingredients that pose challenges for implementation within that approach.

For example, most neural networks use some form of gradient-based (e.g., backpropagation) or Hebbian learning. It has long been argued, however, that backpropagation is not biologically plausible; as \citet{crick89} famously pointed out, backpropagation seems to require that information be transmitted backwards along the axon, which does not fit with realistic models of neuronal function \citep[although recent models circumvent this problem in various ways][]{liao15,scellier16,lillicrap14}. This has not prevented backpropagation being put to good use in connectionist models of cognition or in building deep neural networks for AI.  Neural network researchers must regard it as a very good thing, in this case, that concerns of biological plausibility did not hold back research on this particular algorithmic approach to learning.\footnote{Michael Jordan made this point forcefully in his 2015 speech accepting the Rumelhart Prize.} We strongly agree: Although neuroscientists have not found any mechanisms for implementing backpropagation in the brain, neither have they produced definitive evidence against it. The existing data simply offer little constraint either way, and backpropagation has been of obviously great value in engineering today's best pattern recognition systems. 

Hebbian learning is another case in point.  In the form of long-term potentiation (LTP) and spike-timing dependent plasticity (STDP), Hebbian learning mechanisms are often cited as biologically supported \citep{bi01}. However, the cognitive significance of any biologically grounded form of Hebbian learning is unclear. \citet{gallistel13} have persuasively argued that the critical interstimulus interval for LTP is orders of magnitude smaller than the intervals that are behaviorally relevant in most forms of learning. In fact, experiments that simultaneously manipulate the interstimulus and intertrial intervals demonstrate that no critical interval exists. Behavior can persist for weeks or months, whereas LTP decays to baseline over the course of days \citep{power97}. Learned behavior is rapidly reacquired after extinction \citep{bouton04}, whereas no such facilitation is observed for LTP \citep{dejonge85}. Most relevantly for our focus, it would be especially challenging to try to implement the ingredients described in this article using purely Hebbian mechanisms.

Claims of biological plausibility or implausibility usually rest on rather stylized assumptions about the brain that are wrong in many of their details. Moreover, these claims usually pertain to the cellular and synaptic level, with few connections made to systems level neuroscience and subcortical brain organization \citep{Edelman2015a}. Understanding which details matter and which do not requires a computational theory \citep{Marr1982}. Moreover, in the absence of strong constraints from neuroscience, we can turn the biological argument around: Perhaps a hypothetical biological mechanism should be viewed with skepticism if it is cognitively implausible. In the long run, we are optimistic that neuroscience will eventually place more constraints on theories of intelligence. For now, we believe cognitive plausibility offers a surer foundation. 

\textbf{3. Language is essential for human intelligence.  Why is it not more prominent here?}

We have said little in this article about people's ability to communicate and think in natural language, a distinctively human cognitive capacity where machine capabilities lag strikingly.  Certainly one could argue that language should be included on any short list of key ingredients in human intelligence: for instance, \citet{Mikolov2015} featured language prominently in their recent paper sketching challenge problems and a road map for AI. Moreover, while natural language processing is an active area of research in deep learning \citep[e.g.,][]{Mikolov2013,Bahdanau2015,Xu2015}, it is widely recognized that neural networks are far from implementing human language abilities. The question is, how do we develop machines with a richer capacity for language?

We ourselves believe that understanding language and its role in intelligence goes hand-in-hand with understanding the building blocks discussed in this article. It is also true that language builds on the core abilities for intuitive physics, intuitive psychology, and rapid learning with compositional, causal models that we do focus on. These capacities are in place before children master language, and they provide the building blocks for linguistic meaning and language acquisition \citep{pinker07,ODonnell2015,Xu2007,Kemp2007,jackendoff03,Carey2009}. We hope that by better understanding these earlier ingredients and how to implement and integrate them computationally, we will be better positioned to understand linguistic meaning and acquisition in computational terms, and to explore other ingredients that make human language possible.

What else might we need to add to these core ingredients to get language? Many researchers have speculated about key features of human cognition that gives rise to language and other uniquely human modes of thought: Is it recursion, or some new kind of recursive structure building ability \citep{hauser02,Berwick2016}? Is it the ability to reuse symbols by name \citep{deacon98}? Is it the ability to understand others intentionally and build shared intentionality \citep{tomasello10,Bloom2000,frank09}? Is it some new version of these things, or is it just \emph{more} of the aspects of these capacities that are already present in infants? These are important questions for future work with the potential to expand the list of key ingredients; we did not intend our list to be complete.

Finally, we should keep in mind all the ways that acquiring language extends and enriches the ingredients of cognition we focus on in this article. The intuitive physics and psychology of infants is likely limited to reasoning about objects and agents in their immediate spatial and temporal vicinity, and to their simplest properties and states. But with language, older children become able to reason about a much wider range of physical and psychological situations \citep{Carey2009}. Language also facilitates more powerful learning-to-learn and compositionality \citep{Mikolov2015}, allowing people to learn more quickly and flexibly by representing new concepts and thoughts in relation to existing concepts \citep{Lupyan2015,Lupyan2016}. Ultimately, the full project of building machines that learn and think like humans must have language at its core.

\section{Looking forward}

In the last few decades, AI and machine learning have made remarkable progress: Computer programs beat chess masters; AI systems beat Jeopardy champions; apps recognize photos of your friends; machines rival humans on large-scale object recognition; smart phones recognize (and, to a limited extent, understand) speech. The coming years promise still more exciting AI applications, in areas as varied as self-driving cars, medicine, genetics, drug design and robotics. As a field, AI should be proud of these accomplishments, which have helped move research from academic journals into systems that improve our daily lives.

We should also be mindful of what AI has achieved and what it has not. While the pace of progress has been impressive, natural intelligence is still by far the best example of intelligence. Machine performance may rival or exceed human performance on particular tasks, and algorithms may take inspiration from neuroscience or aspects of psychology, but it does not follow that the algorithm learns or thinks like a person. This is a higher bar worth reaching for, potentially leading to more powerful algorithms while also helping unlock the mysteries of the human mind.

When comparing people and the current best algorithms in AI and machine learning, people learn from less data and generalize in richer and more flexible ways. Even for relatively simple concepts such as handwritten characters, people need to see just one or a few examples of a new concept before being able to recognize new examples, generate new examples, and generate new concepts based on related ones (Figure \ref{fig_characters_challenge}A). So far, these abilities elude even the best deep neural networks for character recognition \citep{Ciresan2012}, which are trained on many examples of each concept and do not flexibly generalize to new tasks. We suggest that the comparative power and flexibility of people's inferences come from the causal and compositional nature of their representations.

We believe that deep learning and other learning paradigms can move closer to human-like learning and thought if they incorporate psychological ingredients including those outlined in this paper. Before closing, we discuss some recent trends that we see as some of the most promising developments in deep learning -- trends we hope will continue and lead to more important advances.

\subsection{Promising directions in deep learning}
\label{sec:directions}

There has been recent interest in integrating psychological ingredients with deep neural networks, especially selective attention \citep{Mnih2014,Bahdanau2015,Xu2015}, augmented working memory \citep{Graves2014,Grefenstette2015,Weston2015,Sukhbaatar2015,Graves2016}, and experience replay \citep{McClelland1995,Mnih2015}. These ingredients are lower-level than the key cognitive ingredients discussed in this paper, yet they suggest a promising trend of using insights from cognitive psychology to improve deep learning, one that may be even furthered by incorporating higher-level cognitive ingredients.

Paralleling the human perceptual apparatus, selective attention forces deep learning models to process raw perceptual data as a series of high-resolution ``foveal glimpses'' rather than all at once. Somewhat surprisingly, the incorporation of attention has led to substantial performance gains in a variety of domains, including in machine translation \citep{Bahdanau2015}, object recognition \citep{Mnih2014}, and image caption generation \citep{Xu2015}. Attention may help these models in several ways. It helps to coordinate complex (often sequential) outputs by attending to only specific aspects of the input, allowing the model to focus on smaller sub-tasks rather than solving an entire problem in one shot. For instance, during caption generation, the attentional window has been shown to track the objects as they are mentioned in the caption, where the network may focus on a boy and then a Frisbee when producing a caption like, ``A boy throws a Frisbee'' \citep{Xu2015}. Attention also allows larger models to be trained without requiring every model parameter to affect every output or action. In generative neural network models, attention has been used to concentrate on generating particular regions of the image rather than the whole image at once \citep{Gregor2014}. This could be a stepping stone towards building more causal generative models in neural networks, such as a neural version of the Bayesian Program Learning model that could be applied to tackling the Characters Challenge (Section \ref{characters_section}).

Researchers are also developing neural networks with ``working memories'' that augment the shorter-term memory provided by unit activation and the longer-term memory provided by the connection weights \citep{Graves2014,Grefenstette2015,Weston2015,Sukhbaatar2015,Reed2015,Graves2016}. These developments are also part of a broader trend towards ``differentiable programming,'' the incorporation of classic  data structures such a random access memory, stacks, and queues, into gradient-based learning systems \citep{DifferentiableProgramming}. For example, the Neural Turing Machine \citep[NTM;][]{Graves2014} and its successor the Differentiable Neural Computer \citep[DNC;][]{Graves2016} are neural networks augmented with a random access external memory with read and write operations that maintains end-to-end differentiability. The NTM has been trained to perform sequence-to-sequence prediction tasks such as sequence copying and sorting, and the DNC has been applied to  solving block puzzles and finding paths between nodes in a graph (after memorizing the graph). Additionally, Neural Programmer-Interpreters learn to represent and execute algorithms such as addition and sorting from fewer examples by observing input-output pairs (like the NTM and DNC) as well as execution traces \citep{Reed2015}. Each model seems to learn genuine programs from examples, albeit in a representation more like assembly language than a high-level programming language.

While this new generation of neural networks has yet to tackle the types of challenge problems introduced in this paper, differentiable programming suggests the intriguing possibility of combining the best of program induction and deep learning. The types of structured representations and model building ingredients discussed in this paper -- objects, forces, agents, causality, and compositionality -- help to explain important facets of human learning and thinking, yet they also bring challenges for performing efficient inference (Section \ref{sec:inference}). Deep learning systems have not yet shown they can work with these representations, but they have demonstrated the surprising effectiveness of gradient descent in large models with high-dimensional parameter spaces.  A synthesis of these approaches, able to perform efficient inference over programs that richly model the causal structure an infant sees in the world, would be a major step forward for building human-like AI 

\begin{figure}
\centering\includegraphics[width=6.5in]{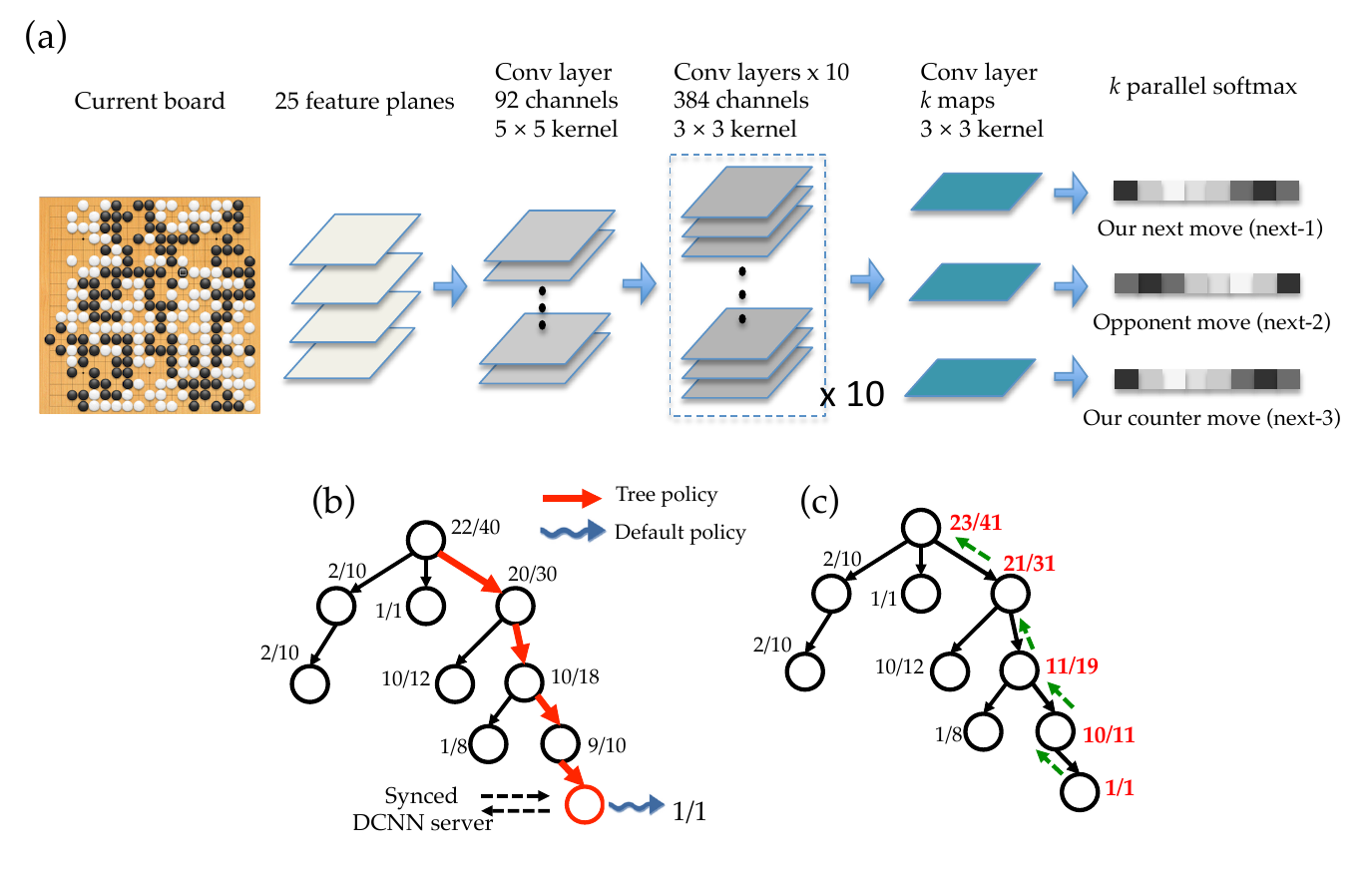}
\caption{An AI system for playing Go combining a deep convolutional network (convnet) and model-based search through Monte-Carlo Tree Search (MCTS). (A) The convnet on its own can be used to predict the next $k$ moves given the current board. (B) A search tree with the current board state as its root and the current ``win/total'' statistics at each node. A new MCTS rollout selects moves along the tree according to the MCTS policy (red arrows) until it reaches a new leaf (red circle), where the next move is chosen by the convnet. From there, play proceeds until the game's end according to a pre-defined default policy based on the Pachi program \citep{Baudivs2012}, itself based on MCTS. (C) The end-game result of the new leaf is used to update the search tree. Adapted from \citet{Tian2015} with permission.}
\label{fig:Go}
\end{figure}

Another example of combining pattern recognition and model-based search comes from recent AI research into the game Go. Go is considerably more difficult for AI than chess, and it was only recently that a computer program -- \textit{AlphaGo} -- first beat a world-class player \citep{alphagonews} by using a combination of deep convolutional neural networks (convnets) and Monte Carlo Tree search \citep{Silver2016}. Each of these components has made gains against artificial and real Go players \citep{Silver2016,Tian2015,Gelly2008,gelly2011monte}, and the notion of combining pattern recognition and model-based search goes back decades in Go and other games.  Showing that these approaches can be integrated to beat a human Go champion is an important AI accomplishment (see Figure \ref{fig:Go}). Just as important, however, are the new questions and directions it opens up for the long-term project of building genuinely human-like AI.

One worthy goal would be to build an AI system that beats a world-class player with the amount and kind of training human champions receive -- rather than overpowering them with Google-scale computational resources. AlphaGo is initially trained on 28.4 million positions and moves from 160,000 unique games played by human experts; it then improves through reinforcement learning, playing 30 million more games against itself. Between the publication of \citet{Silver2016} and before facing world champion Lee Sedol, AlphaGo was iteratively retrained several times in this way; the basic system always learned from 30 million games, but it played against successively stronger versions of itself, effectively learning from 100 million or more games altogether \citep{silverpc}. In contrast, Lee has probably played around 50,000 games in his entire life. Looking at numbers like these, it is impressive that Lee can even compete with AlphaGo at all.  What would it take to build a professional-level Go AI that learns from only 50,000 games? Perhaps a system that combines the advances of AlphaGo with some of the complementary ingredients for intelligence we argue for here would be a route to that end.

AI could also gain much by trying to match the learning speed and flexibility of normal human Go players.  People take a long time to master the game of Go, but as with the Frostbite and Characters challenges (Sections \ref{characters_section} and \ref{frostbite_section}), humans can learn the basics of the game quickly through a combination of explicit instruction, watching others, and experience.  Playing just a few games teaches a human enough to beat someone who has just learned the rules but never played before. Could AlphaGo model these earliest stages of real human learning curves?  Human Go players can also adapt what they have learned to innumerable game variants. The Wikipedia page ``Go variants'' describes versions such as playing on bigger or smaller board sizes (ranging from $9 \times 9$ to $38 \times 38$, not just the usual $19 \times 19$ board), or playing on boards of different shapes and connectivity structures (rectangles, triangles, hexagons, even a map of the English city Milton Keynes).  The board can be a torus, a mobius strip, a cube or a diamond lattice in three dimensions.  Holes can be cut in the board, in regular or irregular ways. The rules can be adapted to what is known as First Capture Go (the first player to capture a stone wins), NoGo (the player who avoids capturing any enemy stones longer wins) or Time Is Money Go (players begin with a fixed amount of time and at the end of the game, the number of seconds remaining on each player's clock is added to their score). Players may receive bonuses for creating certain stone patterns or capturing territory near certain landmarks. There could be four or more players, competing individually or in teams.  In each of these variants, effective play needs to change from the basic game, but a skilled player can adapt and does not simply have to relearn the game from scratch. Could AlphaGo?  While techniques for handling variable sized inputs in convnets may help for playing on different board sizes \citep{Sermanet2013}, the value functions and policies that AlphaGo learns seem unlikely to generalize as flexibly and automatically as people do.  Many of the variants described above would require significant reprogramming and retraining, directed by the smart humans who programmed AlphaGo, not the system itself.  As impressive as AlphaGo is in beating the world's best players at the standard game -- and it is extremely impressive -- the fact that it cannot even conceive of these variants, let alone adapt to them autonomously, is a sign that it does not understand the game as humans do. Human players can understand these variants and adapt to them because they explicitly represent Go \emph{as} a game, with a goal to beat an adversary who is playing to achieve the same goal they are, governed by rules about how stones can be placed on a board and how board positions are scored. Humans represent their strategies as a response to these constraints, such that if the game changes, they can begin to adjust their strategies accordingly.  

In sum, Go presents compelling challenges for AI beyond matching world-class human performance, in trying to match human levels of understanding and generalization, based on the same kinds and amounts of data, explicit instructions, and opportunities for social learning afforded to people. In learning to play Go as quickly and as flexibly as they do, people are drawing on most of the cognitive ingredients this paper has laid out. They are learning-to-learn with compositional knowledge. They are using their core intuitive psychology, and aspects of their intuitive physics (spatial and object representations). And like AlphaGo, they are also integrating model-free pattern recognition with model-based search. We believe that Go AI systems could be built to do all of these things, potentially capturing better how humans learn and understand the game.  We believe it would be richly rewarding for AI and cognitive science to pursue this challenge together, and that such systems could be a compelling testbed for the principles this paper argues for -- as well as building on all of the progress to date that AlphaGo represents.

\subsection{Future applications to practical AI problems} \label{sec:futureproblems}

In this paper, we suggested some ingredients for building computational models with more human-like learning and thought. These principles were explained in the context of the Characters and Frostbite Challenges, with special emphasis on reducing the amount of training data required and facilitating transfer to novel yet related tasks. We also see ways these ingredients can spur progress on core AI problems with practical applications. Here we offer some speculative thoughts on these applications.

\begin{enumerate}

\item \emph{Scene understanding}. Deep learning is moving beyond object recognition and towards scene understanding, as evidenced by a flurry of recent work focused on generating natural language captions for images \citep{Vinyals2014,Karpathy2015,Xu2015}. Yet current algorithms are still better at recognizing objects than understanding scenes, often getting the key objects right but their causal relationships wrong (Figure \ref{caption_generation}). We see compositionality, causality, intuitive physics and intuitive psychology as playing an increasingly important role in reaching true scene understanding. For example, picture a cluttered garage workshop with screw drivers and hammers hanging from the wall, wood pieces and tools stacked precariously on a work desk, and shelving and boxes framing the scene. In order for an autonomous agent to effectively navigate and perform tasks in this environment, the agent would need intuitive physics to properly reason about stability and support. A holistic model of the scene would require the composition of individual object models, glued together by relations. Finally, causality helps infuse the recognition of existing tools (or the learning of new ones) with an understanding of their use, helping to connect different object models in the proper way (e.g., hammering a nail into a wall, or using a saw horse to support a beam being cut by a saw).  If the scene includes people acting or interacting, it will be nearly impossible to understand their actions without thinking about their thoughts, and especially their goals and intentions towards the other objects and agents they believe are present. 

\item \emph{Autonomous agents and intelligent devices}. Robots and personal assistants (such as cellphones) cannot be pre-trained on all possible concepts they may encounter.  Like a child learning the meaning of new words, an intelligent and adaptive system should be able to learn new concepts from a small number of examples, as they are encountered naturally in the environment. Common concept types include new spoken words (names like ``Ban Ki-Moon'' or ``Kofi Annan''), new gestures (a secret handshake or a ``fist bump''), and new activities, and a human-like system would be able to learn to both recognize and produce new instances from a small number of examples. Like with handwritten characters, a system may be able to quickly learn new concepts by constructing them from pre-existing primitive actions, informed by knowledge of the underlying causal process and learning-to-learn.

\item \emph{Autonomous driving}. Perfect autonomous driving requires intuitive psychology. Beyond detecting and avoiding pedestrians, autonomous cars could more accurately predict pedestrian behavior by inferring mental states, including their beliefs (e.g., Do they think it is safe to cross the street? Are they paying attention?) and desires (e.g., Where do they want to go? Do they want to cross? Are they retrieving a ball lost in the street?). Similarly, other drivers on the road have similarly complex mental states underlying their behavior (e.g., Do they want to change lanes? Pass another car? Are they swerving to avoid a hidden hazard? Are they distracted?). This type of psychological reasoning, along with other types of model-based causal and physical reasoning, are likely to be especially valuable in challenging and novel driving circumstances for which there is little relevant training data (e.g. navigating unusual construction zones,  natural disasters, etc.)

\item \emph{Creative design}. Creativity is often thought to be a pinnacle of human intelligence: chefs design new dishes, musicians write new songs, architects design new buildings, and entrepreneurs start new businesses. While we are still far from developing AI systems that can tackle these types of tasks, we see compositionality and causality as central to this goal. Many commonplace acts of creativity are combinatorial, meaning they are unexpected combinations of familiar concepts or ideas \citep{Boden1998,Ward1994}. As illustrated in Figure \ref{fig_characters_challenge}-iv, novel vehicles can be created as a combination of parts from existing vehicles, and similarly novel characters can be constructed from the parts of stylistically similar characters, or familiar characters can be re-conceptualized in novel styles \citep{Rehling2001}. In each case, the free combination of parts is not enough on its own: While compositionality and learning-to-learn can provide the parts for new ideas, causality provides the glue that gives them coherence and purpose.

\end{enumerate}

\subsection{Towards more human-like learning and thinking machines}

Since the birth of AI in the 1950s, people have wanted to build machines that learn and think like people. We hope researchers in AI, machine learning, and cognitive science will accept our challenge problems as a testbed for progress. Rather than just building systems that recognize handwritten characters and play Frostbite or Go as the end result of an asymptotic process, we suggest that deep learning and other computational paradigms should aim to tackle these tasks using as little training data as people need, and also to evaluate models on a range of human-like generalizations beyond the one task the model was trained on. We hope that the ingredients outlined in this article will prove useful for working towards this goal: seeing objects and agents rather than features, building causal models and not just recognizing patterns, recombining representations without needing to retrain, and learning-to-learn rather than starting from scratch.

\subsection*{Acknowledgments}
We are grateful to Peter Battaglia, Matt Botvinick, Y-Lan Boureau, Shimon Edelman, Nando de Freitas, Anatole Gershman, George Kachergis, Leslie Kaelbling, Andrej Karpathy, George Konidaris, Tejas Kulkarni, Tammy Kwan, Michael Littman, Gary Marcus, Kevin Murphy, Steven Pinker, Pat Shafto, David Sontag, Pedro Tsividis, and four anonymous reviewers for helpful comments on early versions of this manuscript. Tom Schaul was very helpful in answering questions regarding the DQN learning curves and Frostbite scoring. This work was supported by the Center for Minds, Brains and Machines (CBMM), under NSF STC award CCF-1231216, and the Moore-Sloan Data Science Environment at NYU.

\bibliographystyle{apacite}
\bibliography{library_clean,library2}
\end{document}